\documentclass[review,11pt]{ReportTemplate}
\usepackage{bm}
\usepackage{graphicx}
\usepackage{paralist,algorithmic,algorithm}
\usepackage{amsmath,mathrsfs,amssymb}

\usepackage[colorlinks,linkcolor=blue]{hyperref} 
\usepackage{url}            
\usepackage{amsfonts}       
\usepackage{nicefrac}       
\usepackage{microtype}      
\usepackage{booktabs}       
\usepackage{longtable}
\usepackage{threeparttable}
\usepackage{multirow}
\usepackage{multicol}
\usepackage{times}
\usepackage{soul}
\usepackage[american]{babel}
\usepackage{algorithm}
\usepackage{algorithmic}
\usepackage{setspace}
\usepackage{graphicx}
\usepackage{subfigure}
\usepackage{wrapfig}
\usepackage{epstopdf}
\usepackage{graphics,epsfig}
\newfloat{figtab}{htb}{fgtb}
\usepackage{stfloats}
\usepackage{siunitx}   
\usepackage{upgreek}   
\usepackage{lineno}    

\begin{document}
	\begin{frontmatter}
		\title{Flexible Transmitter Network}
		
		\author{Shao-Qun Zhang}
		\author{Zhi-Hua Zhou\footnote{Zhi-Hua Zhou is the Corresponding Author.}}
		\address{National Key Laboratory for Novel Software Technology\\
			Nanjing University, Nanjing 210023, China\\~\\
			\normalsize \{zhangsq,zhouzh\}@lamda.nju.edu.cn
		}	
	
\begin{abstract}
Current neural networks are mostly built upon the MP model, which usually formulates the neuron as executing an activation function on the real-valued weighted aggregation of signals received from other neurons. In this paper, we propose the \emph{Flexible Transmitter} (FT) model, a novel bio-plausible neuron model with flexible synaptic plasticity. The FT model employs a pair of parameters to model the transmitters between neurons and puts up a neuron-exclusive variable to record the regulated neurotrophin density, which leads to the formulation of the FT model as a two-variable two-valued function, taking the commonly-used MP neuron model as its special case. This modeling manner makes the FT model not only biologically more realistic, but also capable of handling complicated data, even time series. To exhibit its power and potential, we present the \emph{Flexible Transmitter Network} (FTNet), which is built on the most common fully-connected feed-forward architecture taking the FT model as the basic building block. FTNet allows gradient calculation and can be implemented by an improved back-propagation algorithm in the complex-valued domain. Experiments on a board range of tasks show the superiority of the proposed FTNet. This study provides an alternative basic building block in neural networks and exhibits the feasibility of developing artificial neural networks with neuronal plasticity.
\end{abstract}

\end{frontmatter}

\section{Introduction} \label{sec:intro}
The fundamental computational unit of neural networks is the \emph{neuron}, corresponding to the cell in biological nervous systems. Though neural networks have been studied for more than half a century, and various neural network algorithms and network architectures have been developed, the modeling of neurons is relatively less considered. 

\begin{wrapfigure}{r}{0.4\textwidth}
\centering
\includegraphics[width=0.4\textwidth]{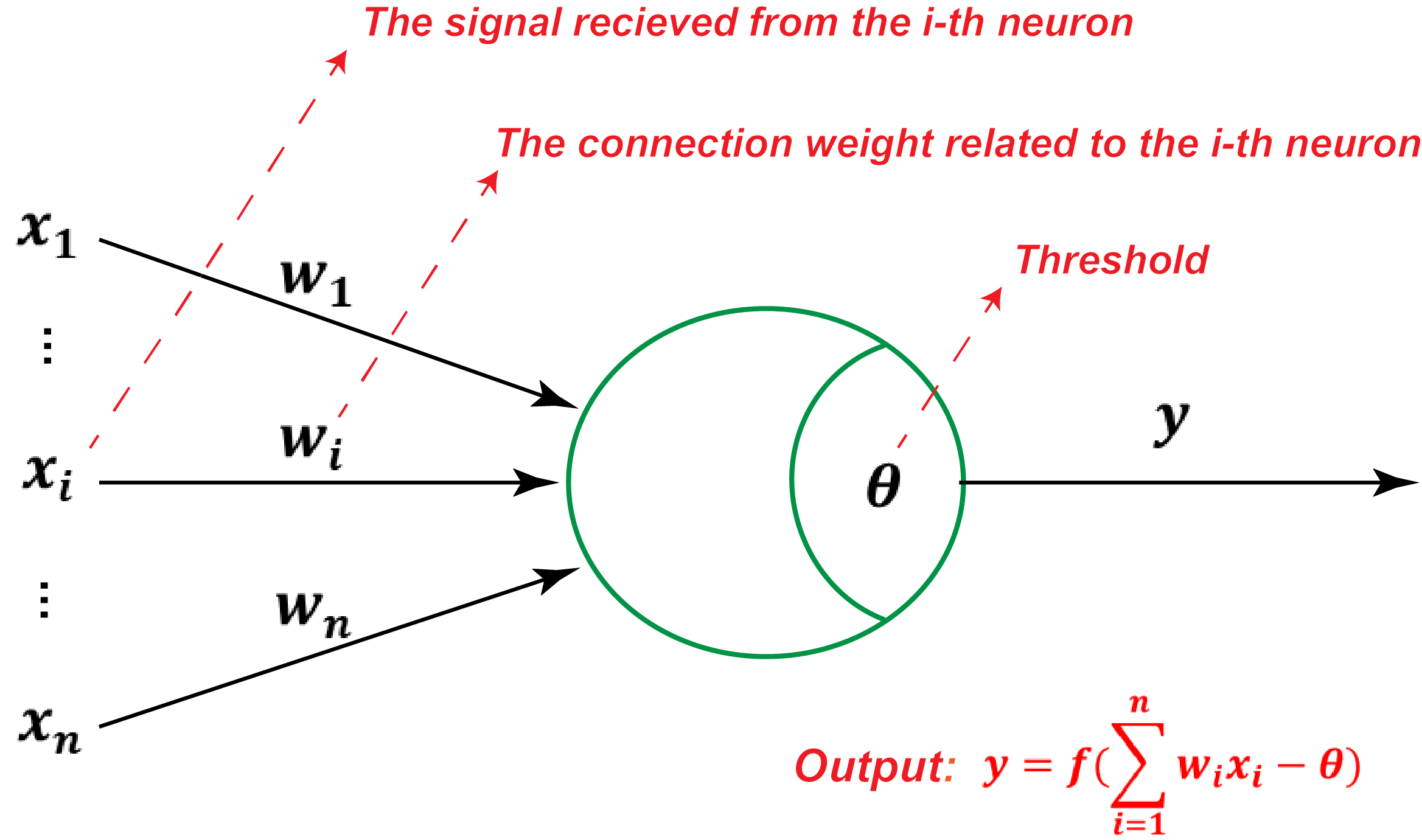}
\caption{The MP model}\label{fig:MP}
\end{wrapfigure}
The most famous and commonly used formulation of neuron is the MP model~\cite{mcculloch1943}, as illustrated in Figure~\ref{fig:MP}. This model formulates the neuron as executing an activation function on the weighted aggregation of signals received from other neurons comparing with a threshold, that is, $y=f(\sum_{i=1}^{n}w_ix_i-\theta)$. In this figure, $x_i$'s are the signals from other neurons, $w_i$'s are the corresponding connection weights, $\theta$ is the threshold of the neuron, and $f$ is the activation function which is usually continuous and differentiable, such as the \emph{sigmoid} function often used in shallow networks and the ReLU function usually used in deep ones.

The MP model is very successful though the formulated cell behavior is quite simple. Real nervous cells are much more complicated, and thus, exploring other bio-plausible formulation with neuronal plasticity is a fundamental problem. There have been many efforts on modeling the spiking behavior of cells, leading to spiking neuron models and pulsed neural networks~\cite{gerstner2002,vanrullen2005}. In this work, we consider another interesting aspect and propose a novel type of neuron model.

\subsection{Synaptic Plasticity}
\begin{figure}[t]
	\centering
	\includegraphics[width=1\linewidth]{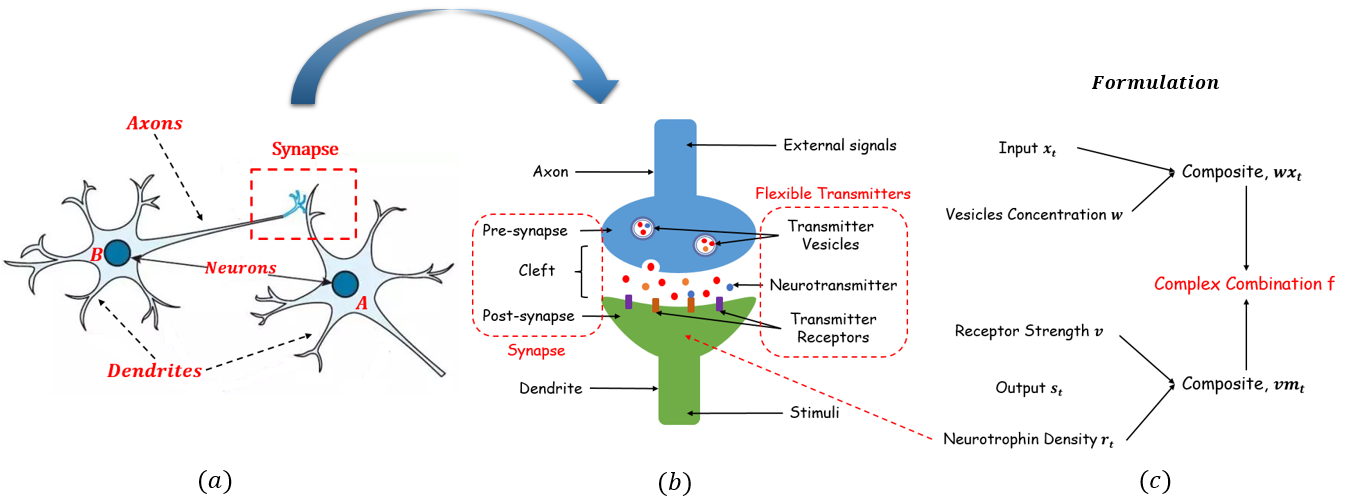}
	\caption{(a) Illustration of biological neurons, (b) synapse with one-way communication, and (c) the modeling of FT model.}  \label{fig:FTNeuron}
\end{figure}

Neuroscience studies~\cite{lodish2008,debanne2011} disclose that the communication between neurons relies on the \emph{synapse}. Signal flows in one direction, from the pre-synaptic neuron to the post-synaptic neuron via the synapse. The synapse usually forms between the endings (or terminals) of the axon and dendrite, which link to the pre-synaptic and post-synaptic neurons, respectively. The endings of the axon and dendrite are named as \emph{pre-synapse} and \emph{post-synapse}, respectively. In common synaptic structures, there is a gap (called \emph{synaptic cleft} in neuroscience) of about \mbox{20 $\upmu$m} between the pre-synapse and post-synapse. The synapse is a combination of the pre-synapse, synaptic cleft, and post-synapse, as shown in the left half of Figure~\ref{fig:FTNeuron}(b).

Synapse ensures the one-way communication behavior between two neurons. In detail, when an external signal through the axon arrives at the pre-synapse, it will be collected by the \emph{synaptic vesicles} and converted into a chemical substance called a \emph{neurotransmitter}~\cite{mattson1988}. With the chemical movement, synaptic vesicles fuse with the pre-synaptic membrane and open channel-like protein molecules, releasing neurotransmitters into the synaptic cleft. Neurotransmitters diffuse across the synaptic cleft, and then bind to the \emph{transmitter receptors} on the post-synapse. The binding chemical action alters the shape and concentration of the transmitter receptors, leading to the opening or closing of ion channels in the cell membrane. Some researchers~\cite{schinder2000,park2013} point out that thanks to these binding chemical actions, the target tissue on post-synapse will secrete a class of proteins, called \emph{neurotrophins}, and the neurotrophin density can alter the tissue size of the synapse, especially the post-synapse. In detail, when the transmitter receptors receive inhibition signals, the neurotrophin density reduces as well as the post-synapse shrinks, and then the shrinking post-synapse inhibits subsequent signal acceptance; while the transmitter receptors receive excitation signals, the neurotrophin density increases as well as the post-synapse swells up, contributing to subsequent signal acceptance~\cite{bi1998}. In summary, regulated by the neurotrophins, the tissue size of the synapses will constantly change, and the generated stimuli will be persistent strengthening or weakening based on recent patterns of neurotrophins, that is, the Long-Term Potentiation or Depression~\cite{fitzsimonds1997,cooke2006}. Finally, the generated stimuli are transmitted to the post-synaptic neuron via the dendrite. The procedure are shown in the right half of Figure~\ref{fig:FTNeuron}(b).

Based on the neurological knowledge about synaptic plasticity, we review the modeling methods of the classical MP and spiking neuron models. The MP model simply takes the whole synapse as a connection parameter (i.e., the $w_i$ in Figure~\ref{fig:MP}), and the signals through this synapse are weighted by $w_i$, leading to its formulation as a real-valued function. The spiking neuron model has established on the \emph{post-synaptic potential} (PSP) assumption, that is, the membrane potential modified by the neurotransmitters will be integrated on the post-synapse, and then the post-synapse is activated only the integrated potential exceeds a threshold. The spiking neuron models usually take the PSP as a variable, and use some first-order differential equations to simulate the potential-integrated process. The common and popular spiking neuron models contain the Integrated-and-Fire model and the Hodgkin-and-Huxley model.

\subsection{Flexible Transmitters}
Back to the one-way neurotransmitter communication mechanism, the signal transmitted from a neuron to another one has undergone several roles, that is, the external signal until being collected by the synaptic vesicles, the neurotransmitters that diffuse across the whole synapse, and the stimuli generated by the post-synapse. And two transformations are carried out around these three roles. One is to convert the external signal to the neurotransmitters, guided by the synaptic vesicles, and the other is to transform the neurotransmitters as the stimuli, which is dominated by the receptors and the neurotrophins of the post-synapse. Therefore, synaptic vesicles, receptors, and neurotrophins are three key elements of the whole communication process. Synaptic vesicles compound with the external signals, and the vesicle concentration makes an effect on the number of neurotransmitters. Receptors are also composite with the neurotransmitters, where the receptor strength controls the amount of the neurotransmitters passing through the membrane. The passed neurotransmitters not only be converted to the stimuli, but also affect the neurotrophin density of the post-synapse, and then the altered neurotrophin density in return affects the passing of the subsequent neurotransmitters. Therefore, the neurotrophin density works like a variable, not only as the output of the neuron model at the current moment but also the input at the subsequent moment, while both the synaptic vesicles and receptors are weighted transmitters. 

Here, we name the combination of the synaptic vesicles and receptors as the \emph{Flexible Transmitter} (FT), as shown in Figure~\ref{fig:FTNeuron}(b), and employs a pair of learnable parameters $(w,v)$ to represent them respectively. Besides, we specifically put up a variable $r_t$ to denote the neurotrophin density at the $t$ time. Obviously, $r_t$ is not only the output of our model at the $t$ time, but also the input at the next time. The chemical action that generates stimuli and alters the neurotrophin density on the post-synapse is believed to be more complex, here we utilize an apposite function $f$ to characterize it. Thereby, a novel neuron model established, which consists of two inputs (i.e., the external signal $x$ and the neurotrophin density at the last time), two outputs (i.e., the generated stimuli $s$ and the the neurotrophin density at the current time), and a pair of learnable parameters $(w,v)$. The FT model has the formation of a two-variable two-valued function. It's entirely different to the conventional neuron models. In Section~\ref{sec:FT}, we will introduce the FT model in detail.

Regard the FT model as the basic building block, various network architectures can be tried, whereas the simplest may be the full-connected feed-forward architecture that has been popularly applied with MP model. In order to demonstrate the power and potential of our FT model, in Section~\ref{sec:FTNet}, we present the \emph{Flexible Transmitter Network} (FTNet), which is a full-connected network constructed by replacing the real-valued MP model with the FT model. Correspondingly, a practicable and effective back-propagation algorithm for training FTNet is developed. Experiments on broad range spatio-temporal data sets have been conducted in Section~\ref{sec:experiment}. The results show that FTNet is able to get excellent performance with the same setting. Finally, we make a discussion in Section~\ref{sec:Discussions} and conclude our work in Section~\ref{sec:conclusion}.

\section{Flexible Transmitter Model} \label{sec:FT}
The interesting discovery of neuroscience in Figure~\ref{fig:FTNeuron}(b) suggests that, the response of neuron A to the external signal from neuron B depends on not only a flexible transmitter structure in the synapse but also the neurotrophin density of the post-synapse.

Inspired by this recognition, we propose the FT model, as illustrated in Figure~\ref{fig:FTNeuron}(c). In contrast to the MP model that the interaction between two neurons is formulated by a single connection weight, in FT model the interaction comprises two parts: $wx_t$ where $x_t$ is the external signal sent to the concerned neuron via the corresponding vesicle concentration $w$, and $v r_{t-1}$ where $r_{t-1}$ is the neurotrophin density at $(t-1)$-th timestamp related to the receptor strength $v$. In other words, the FT model employs a pair of transmitter parameters $(w,v)$ rather than a real-valued weight $w$ in the MP model. Besides, the output of FT neuron at the $t$-th timestamp also consists of two parts: $s_t$ and $r_t$, where $s_t$ is the generated bio-electric/chemical stimulus and $r_t$ indicates the current neurotrophin density of the concerned neuron. After each timestamp, the stimulus signal $s_t$ will be transmitted to the next neuron, while the neurotrophin density will be renewed by $r_t$ in turn and participate in the input of the next moment.

In summary, the proposed FT model employs a pair of parameters $(w,v)$ to indicate the transmitters and puts up an exclusive variable $r_t$ to represent the regulated neurotrophin density. Therefore, the FT model intrinsically has a formation of a two-variable two-valued function $f$ with a pair of parameters $(w,v)$:
\begin{equation} \label{eq:FT}
(s_t,r_t) = f(wx_t,vr_{t-1}).
\end{equation}
We call this model \emph{Flexible Transmitter}.

The FT model has many benefits. Firstly, paired parameters precisely clarifies the roles of the transmitters and provides greater flexibility for synaptic plasticity. From a formulaic point of view, the MP model is a special case of the FT model, when ignoring the transmitter parameter $v$ and the neurotrophin density $r_{t-1}$, or forcing these two values to 0. Secondly, the FT model employs an exclusive variable $r_t$ to indicate the neurotrophin density. During the learning process, the neurotrophin density variable constantly achieves self-renewal, thus deriving a local recurrent system. Therefore, the FT model is able to handle more complicated data, even time series signals.

\section{Flexible Transmitter Network} \label{sec:FTNet}

\subsection{An Implementation of FT Model}
According to Equation~\ref{eq:FT}, the FT model is inherently dominated by a two-variable two-valued function $f$ and a pair of parameters $(w,v)$. Both the input and output of the FT model comprise two parts, and their relationship would be quite complicated since the outputs $s_t$ and $r_t$ share common parameters $(w,v)$ and input $(x_t,r_{t-1})$. Existing neuron models are dependent on one-valued (or real-valued) functions, so the related technologies are hard to be directly applied to this concern. An interesting solution is to resort to a complex-valued formulation that represents the input and output of the concerned neurons, respectively, leading to the neuron model as follows:
\begin{equation} \label{eq:cr}
s_t + r_t \boldsymbol{i} = f( wx_t + vr_{t-1} \boldsymbol{i} ).
\end{equation}

Thanks to complex analysis, the real and imaginary parts of the output of a complex-valued function are geminous twins; both $s_t$ and $r_t$ share the common complex-valued function $f$ and parameters $(w,v)$. So given a specific function $f$, if mastered the value or formulation of $s_t$, we could easily derive $r_t$. Further, once the stimulus $s_t$ is supervised by some teacher signals, even if leaving the neurotrophin density $r_t$ unsupervised, $r_t$ can still be corrected according to the twins' law.

Finally, we emphasize that using complex-valued function is just one kind of approach for implementing the proposed FT model. It may not be the most appropriate one, and it is very likely that there are better approaches to be explored in the future.

\subsection{A Simple Architecture of FT Net}
The FT neuron is a fundamental unit of neural networks. To evaluate its potential, we consider to use the simplest full-connected feed-forward neural network through replacing the common MP model by the FT model as its basic building block, thus we get the FTNet. Based on Equation~\ref{eq:cr}, we can provide a general vectorized representation for a layer of FT neurons:
\begin{equation} \label{eq:cr_layer}
\boldsymbol{s}_t + \boldsymbol{r}_t \boldsymbol{i} = f(\mathbf{W}\boldsymbol{x}_t + \mathbf{V}\boldsymbol{r}_{t-1} \boldsymbol{i}).
\end{equation}
It is worth noting that if considering $m$-dimensional external input signals $\boldsymbol{x}_t$ and $n$-dimensional output stimuli $\boldsymbol{s}_t$, then the transmitter concentration matrices $\mathbf{W} \in \mathbb{R}^{n \times m}$, $\mathbf{V} \in \mathbb{R}^{n \times n}$ and the neurotrophin density vectors $\boldsymbol{r}_t$ and $\boldsymbol{r}_{t-1}$ are $n$-dimensional. Reusing the layer-vectorized representation in Equation~\ref{eq:cr_layer} layer by layer, we can obtain a multi-layer full-connected feed-forward architecture of FTNet.

There remains two unsolved problems: (1) what is the complex-valued function $f$ like? and (2) how to train it? To address these problems, we divide the complex-valued function $f$ in Equation~\ref{eq:cr} into two parts: a conversion function $\tau:\mathbb{C} \rightarrow \mathbb{C}$ and an activation function $\sigma:\mathbb{C} \rightarrow \mathbb{C}$, where $f = \sigma \circ \tau$. This composition operation separates the complex structure in $f$ from the nonlinear activation; the conversion function $\tau$ formulates the complex aggregation, usually differentiable, while $\sigma$ puts stress on the formulation of activations. Thus, FTNet allows gradient calculation and adapts to some conventional activation functions.

In order to perform back-propagation in a FTNet, it's necessary for the conversion function to be differentiable, i.e., holomorphic. The detailed information about holomorphism can be obtained in Appendix~\ref{app:A}. By the holomorphism, we restricted the set of possible conversion functions. Nevertheless, there are still various holomorphic functions can be tried. Ideally, we can get inspiration from bio-science to design this conversion function; here we have not incorporated bio-knowledge and simple use the simplest linear holomorphic function as follows:
\begin{equation}  \label{eq:holomorphic}
\begin{aligned}
\tau( \mathbf{W} \boldsymbol{x}_t + \mathbf{V} \boldsymbol{r}_{t-1} ) & =( \mathbf{W} \boldsymbol{x}_t + \mathbf{V} \boldsymbol{r}_{t-1} \boldsymbol{i} ) \cdot ( a + b \boldsymbol{i} ) \\
& = ( a \mathbf{W} \boldsymbol{x}_t - b \mathbf{V} \boldsymbol{r}_{t-1} ) + ( b \mathbf{W} \boldsymbol{x}_t + a \mathbf{V} \boldsymbol{r}_{t-1} ) \boldsymbol{i},
\end{aligned}
\end{equation}
where $a$ and $b$ are constants in $\mathbb{R}$. Then Equation~\ref{eq:cr_layer} becomes:
\begin{equation}  \label{eq:activation}
\boldsymbol{s}_t + \boldsymbol{r}_t \boldsymbol{i} = \sigma \left(~ ( a \mathbf{W} \boldsymbol{x}_t - b \mathbf{V} \boldsymbol{r}_{t-1} ) + ( b \mathbf{W} \boldsymbol{x}_t + a \mathbf{V} \boldsymbol{r}_{t-1}) \boldsymbol{i} ~\right).
\end{equation}

Next, we are going to introduce some activation functions that can generally be used in FTNet. An intuitive idea is to decompose the activation function $\sigma$ into two real-valued nonlinear functions that are activated with respect to the real and imaginary parts, respectively, that is, $\sigma  = \sigma_{real} + \sigma_{imag} \boldsymbol{i}$, where $\sigma_{real}$ and $\sigma_{imag}$ are real-valued activation functions, such as the \emph{sigmoid} and \emph{tanh} functions. Apart from this, FTNet also allow non-trivial activations in the complex-valued domain, such as the modReLU~\cite{arjovsky2016} and zReLU~\cite{trabelsi2018}. 

Finally, by employing the holomorphic conversion and complex-valued activation functions, a complete FTNet is established and we call this implementation procedure the \emph{Complex-valued Reaction}. For a $L$-layer FTNet, and its feed-forward procedure runs as follows:
\begin{equation}
	\left\{\begin{aligned}
	& \boldsymbol{s}^0_t = \boldsymbol{x}_t, \\
	& \boldsymbol{\alpha}^l_t = a \mathbf{W}^l \boldsymbol{s}^{l-1}_t - b \mathbf{V}^l \boldsymbol{r}_{t-1}, \\
	& \boldsymbol{\beta}^l_t = b \mathbf{W}^l \boldsymbol{s}^{l-1}_t + a \mathbf{V}^l \boldsymbol{r}_{t-1}, \\
	& \boldsymbol{s}^l_t + \boldsymbol{r}^l_t \boldsymbol{i} = \sigma ( \boldsymbol{\alpha}^l_t + \boldsymbol{\beta}\boldsymbol{i}^l_t ), \\
	& \boldsymbol{y}_t = \boldsymbol{s}_t^L. \\
	\end{aligned}\right.
\end{equation}
Throughout this paper, we will use the notations FT0 to denote a one-layer FTNet, i.e., without any hidden layer, and FT1 to indicate the FTNet with only one hidden layer. The cascade structures of FT0 and FT1 are abbreviated as size$(m,0,n)$ and size$(m,l,n)$, respectively, where $l$ is the number of hidden neurons.

\subsection{Complex Back-Propagation}
We present the \emph{Complex Back-Propagation} (CBP) algorithm for training the FTNet. CBP is an extension of the common back-propagation algorithm in the complex-valued domain. The core idea of CBP is to take the neurotrophin density as an implicit variable, so that the desired gradients become a function of the partial derivative of $r_t$ with respect to the connection parameters $\mathbf{W}$ and $\mathbf{V}$. Here we just list the main steps and results of our proposed CBP algorithm, and a concrete introduction is detailed in Appendix~\ref{app:CBP}. 

Considering the loss function for FTNet in time interval $[0,T]$,
\[
\begin{aligned}
\mathbf{E}(\mathcal{W},\mathcal{V}) &= \frac{1}{2} \int_{t=0}^T E_t ~dt \\
& = \frac{1}{2} \int_{t=0}^T \sum_{i=1}^{n_L} ~\left( \hat{\boldsymbol{y}_t}(i) - \boldsymbol{y}_t(i) \right)^2 ~dt,
\end{aligned}
\]
where $\hat{\boldsymbol{y}_t}$ is the supervised signals. Taking into account the temporal dependency, the back-propagation gradients of transmitter concentration matrices through time can be calculated by:
\[
( \mathbf{\nabla_{W^l} E}, \mathbf{\nabla_{V^l} E} ) = \int_{t=0}^T \left( \mathbf{\nabla_{W^l} E}_t, \mathbf{\nabla_{V^l} E}_t \right) ~dt,
\]
and
\begin{equation} \label{eq:nabla}
\left\{\begin{aligned}
\mathbf{\nabla_{W^l} E}_t &= \delta \boldsymbol{s}_t^l ~\frac{\partial~ \boldsymbol{s}_t^l}{\partial~ \boldsymbol{\alpha}^l_t} ~\frac{\partial~ \boldsymbol{\alpha}^l_t}{\partial~ \mathbf{W}^l} ,\\
\mathbf{\nabla_{V^l} E}_t &= \delta \boldsymbol{s}_t^l ~\frac{\partial~ \boldsymbol{s}_t^l}{\partial~ \boldsymbol{\alpha}^l_t} ~\frac{\partial~ \boldsymbol{\alpha}^l_t}{\partial~ \mathbf{V}^l} .
\end{aligned}\right.
\end{equation}
The aforementioned formula consists of three terms. (1) The first term $\delta \boldsymbol{s}_t^l$ is the back-propagation error correction in the $l$-th layer at time $t$. So we can calculate it by $\delta \boldsymbol{s}_t^l = a~ (\mathbf{W}^{l+1})^T~ \left( \delta \boldsymbol{s}_t^{l+1} \odot \sigma'(\boldsymbol{\alpha}_t^{l+1} + \boldsymbol{\beta}_t^{t+1} \boldsymbol{i}) \right)$, where $\odot$ is the point-wise operation. (2) The second term $\frac{\partial~ \boldsymbol{s}_t^l}{\partial~ \boldsymbol{\alpha}^l_t}$ is a diagonal matrix, where its diagonal elements are the point-wise derivatives of activation $\sigma$. (3) The third term $\frac{\partial~ \boldsymbol{\alpha}^l_t}{\partial~ \mathbf{W}^l}$ and $\frac{\partial~ \boldsymbol{\alpha}^l_t}{\partial~ \mathbf{V}^l}$ are tensors, belonging to $\mathbb{R}^{T \times n_l \times n_{l-1} }$. So Equation~\ref{eq:nabla} can be detailed as follows:
\[
\left\{\begin{aligned}
\mathbf{\nabla_{W^l} E}_t &= \left( \delta \boldsymbol{s}_t^l (1) ~\frac{\partial~ \boldsymbol{s}_t^l(1)}{\partial~ \boldsymbol{\alpha}^l_t(1)} ~\frac{\partial~ \boldsymbol{\alpha}^l_t(1)}{\partial~ \mathbf{W}^l}, \cdots, \delta \boldsymbol{s}_t^l (n_l) ~\frac{\partial~ \boldsymbol{s}_t^l(n_l)}{\partial~ \boldsymbol{\alpha}^l_t(n_l)} ~\frac{\partial~ \boldsymbol{\alpha}^l_t(n_l)}{\partial~ \mathbf{W}^l} \right) ,\\
\mathbf{\nabla_{V^l} E}_t &= \left( \delta \boldsymbol{s}_t^l (1) ~\frac{\partial~ \boldsymbol{s}_t^l(1)}{\partial~ \boldsymbol{\alpha}^l_t(1)} ~\frac{\partial~ \boldsymbol{\alpha}^l_t(1)}{\partial~ \mathbf{V}^l}, \cdots, \delta \boldsymbol{s}_t^l (n_l) ~\frac{\partial~ \boldsymbol{s}_t^l(n_l)}{\partial~ \boldsymbol{\alpha}^l_t(n_l)} ~\frac{\partial~ \boldsymbol{\alpha}^l_t(n_l)}{\partial~ \mathbf{V}^l} \right) .
\end{aligned}\right.
\]
$\left( \frac{\partial~ \boldsymbol{\alpha}^l_t}{\partial~ \mathbf{W}^l}, \frac{\partial~ \boldsymbol{\alpha}^l_t}{\partial~ \mathbf{V}^l} \right)$ are the core of our CBP algorithm, including two back-propagation pipelines with respect to $\mathbf{W}^l$ and $\mathbf{V}^l$, respectively. Regarding the neurotrophin density as an implicit variable, these tensors can be calculated as follows: 
\[
\left\{\begin{aligned}
\frac{\partial~ \boldsymbol{\alpha}^l_t(i)}{\partial~ \mathbf{W}^l(j,k)} &= \left\{\begin{aligned}
& a \boldsymbol{s}_t^{l-1}(k) - b \left( \sum_{h=1}^{n_l} \mathbf{V}^l(i,h) ~\frac{\partial~ \boldsymbol{r}_{t-1}^l(h)}{\partial~ \mathbf{W}^l(i,k)} \right) &,~~ j=i ,\\
& - b \left( \sum_{h=1}^{n_l} \mathbf{V}^l(j,h) ~\frac{\partial~ \boldsymbol{r}_{t-1}^l(h)}{\partial~ \mathbf{W}^l(j,k)} \right) &,~~ j \neq i ,\\
\end{aligned}\right.  \\
\frac{\partial~ \boldsymbol{\alpha}^l_t(i)}{\partial~ \mathbf{V}^l(j,h)} &=
\left\{\begin{aligned}
& -b \left( \boldsymbol{r}_{t-1}^l(h) + \sum_{h=1}^{n_l} \mathbf{V}^l(i,h) ~\frac{\partial~ \boldsymbol{r}_{t-1}^l(h)}{\partial~ \mathbf{V}^l(i,h)} \right)  &,~~ j=i ,\\
& -b \left( \sum_{h=1}^{n_l} \mathbf{V}^l(j,h) ~\frac{\partial~ \boldsymbol{r}_{t-1}^l(h)}{\partial~ \mathbf{V}^l(j,h)} \right) &,~~ j \neq i .\\
\end{aligned}\right.  \\
\end{aligned}\right.
\]
In the feed-forward process, the stimulus signals and neurotrophin densities are outputted by holomorphic FT neurons with coupling processing capability; while in the CBP process, the neurotrophin densities are indirectly regulated by the supervised stimulus signals. So we still need to supply the feed-forward errors caused by the neurotrophin densities. The calculation procedure of partial derivatives $\left( \frac{\partial~ \boldsymbol{r}^l_t}{\partial~ \mathbf{W}^l}, \frac{\partial~ \boldsymbol{r}^l_t}{\partial~ \mathbf{V}^l} \right)$ is similar to that of $\left( \frac{\partial~ \boldsymbol{\alpha}^l_t}{\partial~ \mathbf{W}^l}, \frac{\partial~ \boldsymbol{\alpha}^l_t}{\partial~ \mathbf{V}^l} \right)$. Therefore, we directly provides the results:
\[
\left\{\begin{aligned}
\frac{\partial~ \boldsymbol{r}^l_t(i)}{\partial~ \mathbf{W}^l(j,k)} &= \frac{\partial~ \boldsymbol{r}^l_t(i)}{\partial~ \boldsymbol{\beta}^l(i)} ~\frac{\partial~ \boldsymbol{\beta}^l_t(i)}{\partial~ \mathbf{W}^l(j,k)}, \\
\frac{\partial~ \boldsymbol{r}^l_t(i)}{\partial~ \mathbf{V}^l(j,h)} &= \frac{\partial~ \boldsymbol{r}^l_t(i)}{\partial~ \boldsymbol{\beta}^l(i)} ~\frac{\partial~ \boldsymbol{\beta}^l_t(i)}{\partial~ \mathbf{V}^l(j,h)}, \\
\end{aligned}\right.
\]
and 
\[
\left\{\begin{aligned}
\frac{\partial~ \boldsymbol{\beta}^l_t(i)}{\partial~ \mathbf{W}^l(j,k)} &= \left\{\begin{aligned}
& b \boldsymbol{s}_t^{l-1}(k) + a \left( \sum_{h=1}^{n_l} \mathbf{V}^l(i,h) ~\frac{\partial~ \boldsymbol{r}_{t-1}^l(h)}{\partial~ \mathbf{W}^l(i,k)} \right) &,~~ j=i ,\\
& a \left( \sum_{h=1}^{n_l} \mathbf{V}^l(j,h) ~\frac{\partial~ \boldsymbol{r}_{t-1}^l(h)}{\partial~ \mathbf{W}^l(j,k)} \right) &,~~ j \neq i ,\\
\end{aligned}\right.   \\
\frac{\partial~ \boldsymbol{\beta}^l_t(i)}{\partial~ \mathbf{V}^l(j,h)} &=
\left\{\begin{aligned}
& a \left( \boldsymbol{r}_{t-1}^l(h) + \sum_{h=1}^{n_l} \mathbf{V}^l(i,h) ~\frac{\partial~ \boldsymbol{r}_{t-1}^l(h)}{\partial~ \mathbf{V}^l(i,h)} \right) &,~~ j=i ,\\
& a \left( \sum_{h=1}^{n_l} \mathbf{V}^l(j,h) ~\frac{\partial~ \boldsymbol{r}_{t-1}^l(h)}{\partial~ \mathbf{V}^l(j,h)} \right) &,~~ j \neq i .\\
\end{aligned}\right.   \\
\end{aligned}\right.
\]
On the basis of the calculation steps, we can obtain the gradients $\left( \mathbf{\nabla_{W^l} E}, \mathbf{\nabla_{V^l} E} \right)$ and correct the concentration parameters according to:
\[
\left\{\begin{aligned}
\hat{\mathbf{W}^l} &= \mathbf{W}^l - \eta \mathbf{\nabla_{W^l} E}, \\
\hat{\mathbf{V}^l} &= \mathbf{V}^l - \eta \mathbf{\nabla_{V^l} E}, \\
\end{aligned}\right.
\]
where $\eta$ is the learning rate.

Before prediction (in the time interval $[0,T]$), we still need to update the imaginary parts (neurotrophin densities $\boldsymbol{r}_{0}^l, \cdots, \boldsymbol{r}_{T}^l$ ) as follows:
\[
\left\{\begin{aligned}
& \boldsymbol{s}^0_t = \boldsymbol{x}_t, \\
& \boldsymbol{r}_t^l = \sigma\left( b \hat{\mathbf{W}^l} \boldsymbol{s}_t^{l-1} + a \hat{\mathbf{V}^l} \boldsymbol{r}_{t-1}^l \right), \\
\end{aligned}\right.
\]
and reset the imaginary errors $\left( \frac{\partial~ \boldsymbol{r}^l_T}{\partial~ \mathbf{W}^l}, \frac{\partial~ \boldsymbol{r}^l_T}{\partial~ \mathbf{V}^l} \right)$ as zeros.

\section{Experiments}  \label{sec:experiment}
In this section, we compare FTNet with several popular neural networks on three data sets. The goal is to demonstrate its superior performance. 

For the configurations of FTNet, we employ $tanh$ as the activation function and 0.01 for learning rate, thanks to the comparsion experiments about activations in Appendix~\ref{app:simluated}. Other hyper-parameters cannot be fixed across tasks, otherwise, the performance will be embarrassingly unsatisfactory. Therefore, we examine a variety of configurations on the testing data set and pick out the one with the best performance. Supplemental materials about these experiments, especially the data sets and configurations are detailed in Appendix~\ref{app:experiments}. Here, we put our focus on the experimental results.

\subsection{Univariate Time Series Forecasting - Yancheng Automobile Registration} \label{subsec:univariate}
We first conduct experiments on the \emph{Yancheng Automobile Registration Forecasting} competition, which is a real-world univariate time series forecasting task. This task is hard since not only objective automobile registration records is the mixture of 5 car brands, but also there exists lots of missing data and sudden changes caused by holiday or other guiding factors which we cannot obtain in advance.

\begin{table}[!htb]
\centering
\caption{MSE and settings of comparative models for the task of forecasting Yancheng Automobile Registration records.}
\label{tab:univariate}
\begin{tabular}{@{}ccccc@{}}
	\toprule
	Types & Models & Settings & Epochs & MSE ($10^5$) \\ 
	\midrule
	\multirow{4}{*}{Statistical Models} 
	& ARIMA                 & $(p,d,q)=(6,1,3)$ & -- & 84.5129\\
	& MAR~\cite{won2013}    & -- & -- & 92.6458 \\
	& AGP~\cite{blaauw2011} & -- & -- & 41.0147 \\
	& KNNs~\cite{yang2011}  & $(K,w)=(1,1)$ & -- & 31.2573\\
	\midrule
	\multirow{5}{*}{Neural Networks} 
	& NARXnet~\cite{guzman2017}  & size(7,10,1) & 80 & 28.0549 \\
	& RNN~\cite{gers2000}        & size(5,50,1) & 100 & 22.2296 \\
	& LSTM~\cite{hochreiter1997} & size(5,50,1) & 80 & 15.2490 \\
	& GRU~\cite{cho2014}  & size(5,50,1) & 85 & 13.0421 \\
	& LSTNet~\cite{lai2018} & size(7,50,1) & 100 & 10.2070 \\
	\midrule
	\multirow{2}{*}{Our Works} 
	& FT0    & size(5,0,1)  & 100  & 23.6634 \\
	& FT1    & size(5,50,1) & 100  & \textbf{4.5067} \\
	\bottomrule
\end{tabular}
\end{table}
We compare our proposed FTNet with several state-of-the-art statistical models and neural networks~\cite{cheng2015}, and evaluate the performance by \emph{Mean Square Error} (MSE). The results are summarized in Table~\ref{tab:univariate}, showing that FT1 achieves highly competitive performance.

\subsection{Multivariate Time Series Forecasting - Traffic Prediction on HDUK}
We also validate FTNet on the \emph{Highway Data of United Kingdom} (HDUK), which is a representative multivariate traffic prediction data set. For convenience, we choose roads with relatively large several traffic flow for study and collect the traffic data of the 12 months in 2011 where the first 10 months are divided as the training set and the later 2 months as the testing set. We also set that input (feature vectors) in this experiment is the normalized value (Total Traffic Flow \& Travel Time \& Fused Average Speed \& Link Length) of all observation points in previous 8 time intervals. Output is the prediction value (Total Traffic Flow) in the next time interval of a target observation point for prediction. Empirically, we add the evaluation indicator, \emph{Confusion Accuracy}, which consists of \emph{True Positive Rate} (TPR) and \emph{True Negative Rate} (TNR), to compare the performance of FTNet with other competing methods. 

\begin{table}[!htb]
\centering
\caption{MSE and confusion accuracy of comparative models for the task of forecasting HDUK.}
\label{tab:traffic}
\begin{threeparttable}
	\begin{tabular}{@{}cccccccc@{}}
		\toprule
		\multicolumn{2}{c}{Models \& Settings} & NARXnet & RNN & LSTM  &LSTNet & FT0 & FT1\\
		data sets & Evaluation(\%) & $\Box$ & $\bigtriangleup$ & $\bigtriangleup$ & $\bigtriangleup$ & $\Diamond$ & $\bigtriangleup$ \\
		\midrule
		\midrule
		\multirow{3}{*}{A1}
		& MSE & 0.0469 & 0.1499 & 0.0262 & 0.0247 & 0.1169 & \textbf{0.0221}\\
		& TPR & 97.20  & 97.20  & 98.13  & 98.13 & 96.20 & \textbf{99.07}\\
		& TNR & 95.29  & 91.74  & 96.47  & 97.41 & 94.12 & \textbf{97.65}\\
		\midrule
		\midrule
		\multirow{3}{*}{A1033} 
		& MSE & 0.1584 & 0.1716 & 0.1397 & 0.1401 & 0.1372 & \textbf{0.1119}\\
		& TPR & 88.51  & 93.10  & 94.25  & 94.11  & 94.11 & \textbf{96.55} \\
		& TNR & 91.43  & 93.33  & 92.38  & 92.25  & 92.38 & \textbf{97.14} \\
		\midrule
		\midrule
		\multirow{3}{*}{A11} 
		& MSE & 0.1754 & 0.1770 & 0.1725 & 0.1690 & 0.1755 & \textbf{0.1651}\\
		& TPR & 97.06  & 96.08  & 97.06  & 97.06 & 97.06 & \textbf{99.02} \\
		& TNR & 95.56  & 91.11  & 93.33  & 95.93 & \textbf{96.67} & 94.44 \\
		\bottomrule
	\end{tabular}
	\begin{tablenotes}
		\item[$\Box$] denotes a size(*,0,1) cascade structure and iterates 100 times;
		\item[$\bigtriangleup$] denotes size(*,100,1) cascade structure and iterates 100 times;
		\item[$\Diamond$] indicates a network configuration with 100 recurrent neurons and 32-dimensional convolution layer and iterates 100 epochs.
	\end{tablenotes}
\end{threeparttable}
\end{table}
Table~\ref{tab:traffic} lists the comparative results of FTNet and other neural networks on a part of HDUK data sets, including the Total Traffic Flow of first 3 crossroads: A1, A1033, and A11. FTNet achieves the best performance in the same settings as other competing neural networks.

\subsection{Image Recognition on Pixel-by-pixel MNIST}
We also conduct the experiments on a benchmark image recognition task, to evaluate the ability of FTNet for processing with spatial information. Pixel-by-pixel MNIST is a popular challenging image recognition data set, which is standard benchmark to test the performance of a learning algorithm~\cite{trabelsi2018}. Following a similar setup to~\cite{pillow2005}, the handwritten digit image is converted into a pixel-by-pixel spiking train with $T=784$. Standard split of 60,000 training samples and 10,000 testing samples was used with no data augmentation. We also add a convolution filter to the external input signals at each time stamp. The size of the convolution kernel is preset as $2 \times 2$.

In this experiment, we also compare our FTNet with another bio-inspired neural network, the spiking neural network (SNN). For classification, we use the spiking counting strategy. In other words, during training, we specify a target of 20 spikes for the true neuron and 5 spikes for each false neuron; while testing, the output class is the one which generates the highest spike count. All models except SLAYER~\cite{shrestha2018} employ a \emph{softmax} function for classification and are optimized by a cross-entropy loss.

\begin{table}[!htb]
\centering
\caption{Accuracy of comparative models for the task of classifying pixel-pixel MNIST.}
\label{tab:MNIST}
\begin{tabular}{@{}cccc@{}}
	\toprule
	Models & Cascade & Epochs & Accuracy (\%) \\ 
	\midrule
	CNN-SVM~\cite{niu2012}     &  --  & --  & 98.79  \\
	SLAYER                     & (*,500,10) & 1000 & 96.93 \\
	uRNN~\cite{arjovsky2016}   & (*,150,10) & 1000 & 97.28 \\
	CNN-RNN                    & (*,150,10) & 800  & 95.21 \\
	CNN-LSTM                   & (*,150,10) & 800  & 98.66 \\
	FT0 (our work)  & (*,0,10) & 1000 & 92.87 \\
	FT1 (our work)  & (*,150,10) & 1000 & \textbf{99.12} \\
	\bottomrule
\end{tabular}
\end{table}
In order to accurately compare the performance of various deep learning models, we force all networks to adopt the same setting (including 150 hidden neurons and the number of iterations), except for SLAYER which is allowed more neurons to ensure convergence. The experimental results are shown in Table~\ref{tab:MNIST}, confirming the superiority of our FTNet to the existing state-of-the-art neural networks.

\section{Discussions} \label{sec:Discussions}

\subsection{About Complex-valued Reaction}
There have been many efforts on developing neural networks using complex-valued formation. For example, \cite{arjovsky2016} forces the connection weights as a unitary matrix in the complex-valued domain for circumventing the issue of vanishing and exploding gradients in RNNs. \cite{trabelsi2018} proposed a complex-valued connection matrix that works like a real-valued 2D convolution for deep neural networks, and this technology later was developed to a quaternion-valued formation~\cite{parcollet2019}.

Our proposed FT model is essentially different from the aforementioned works. (1) The motivations are different. Existing neural networks relative to complex-valued formation are motivated to explore an atomic component for overcoming the drawbacks or improving the representational capacity of neural networks. However, the FT model is a novel type of neuron model, which depicts the neurotransmitter communication mechanism in synaptic plasticity and is formulated by a two-variable two-valued function. The complex-valued reaction is just a valid implementation for the two-variable two-valued function as well as the FT model. (2) The use of complex-valued formation is different. The connection matrices in~\cite{arjovsky2016} are asked to have complex-valued eigenvalues with absolute value 1. Thus the gradients of recurrent networks are guaranteed to avoid explosion. The inputs of \cite{trabelsi2018,parcollet2019} are pre-processed into a complex-valued or quaternion-valued formation for facilitating subsequent convolution-like operation. In this work, we employ a new-born variable, the neurotrophin density to model the behavior that the tissue size of synapses would change as the learning process. For the complex-valued reaction, the neurotrophin densities are regarded as the imaginary parts of the complex-valued function, leading to a local recurrent system in the FT model. The experimental results show that the FT model is with potential.

Finally, we have to note that the complex-valued reaction is just a formulation of the FT model, and many valid implementation approaches are worthy to be tried. Furthermore, numerous holomorphic conversion functions and activations are worthy of being explored; custom-built conversion functions may extract potential and significant adjoint features on some real-world applications. Besides, we here only provided a simplest fully-connected feed-forward network, i.e., the FTNet. Various alternative network architectures can be explored in the future.

\subsection{Comparsion with RNNs}
As mentioned above, the proposed FT model derives a local recurrent system, thus able to handle spatio-temporal signals. Speaking of this, it is easy to remind us of the typical recurrent neural networks. Different from the RNN's unit formalized by a real-valued function $s_t = g(x_t, s_{t-1};w,v)$, the FT model is dominated by $(s_t,r_t) = f(x_t,r_{t-1};w,v)$. Obviously, the RNN's unit is a special case of the FT model; these two models are equivalent to each other when pre-setting $r_t = s_t$. Therefore, with the same number of parameters $w$ and $v$, the FT model has a broader representational capacity.

\section{Conclusion}  \label{sec:conclusion}
In this paper, we proposed the FT model, which is a brand-new model for bio-plausible nervous system. In contrast to the traditional MP model that simply regards the whole nervous synapse as a real-valued parameter, the FT model meticulously depicts the neurotransmitter communication mechanism in synaptic plasticity, that is, employs a pair of parameters to model the transmitters and puts up a variable to denote the regulated neurotrophin density, thus leading to a formulation of a two-variable two-valued function. The FT model takes the MP model and RNN's unit as its special cases. To demonstrate the power and potential of our proposed FT model, we present the FTNet using the most common full-connected feed-forward architecture. For simplicity, we employ the holomorphic complex-valued reaction as an implementation paradigm, and then, present a practicable and effective CBP algorithm for training a FTNet. The experiments conducted on wide-range tasks confirm the effectiveness and superiority of our model.


\newpage
\bibliography{reference}

\begin{thebibliography}{10}

\bibitem{arjovsky2016}
Martin Arjovsky, Amar Shah, and Yoshua Bengio.
\newblock Unitary evolution recurrent neural networks.
\newblock In {\em Proceedings of the 33nd International Conference on Machine
  Learning (ICML)}, pages 1120--1128, 2016.

\bibitem{bi1998}
Guo-qiang Bi and Mu-ming Poo.
\newblock Synaptic modifications in cultured hippocampal neurons: Dependence on
  spike timing, synaptic strength, and postsynaptic cell type.
\newblock {\em Journal of Neuroscience}, 18(24):10464--10472, 1998.

\bibitem{blaauw2011}
Maarten Blaauw, J~Andr{\'e}s Christen, et~al.
\newblock Flexible paleoclimate age-depth models using an autoregressive gamma
  process.
\newblock {\em Bayesian Analysis}, 6(3):457--474, 2011.

\bibitem{cheng2015}
Changqing Cheng, Akkarapol Sa-Ngasoongsong, Omer Beyca, Trung Le, Hui Yang,
  Zhenyu Kong, and Satish~TS Bukkapatnam.
\newblock Time series forecasting for nonlinear and non-stationary processes: A
  review and comparative study.
\newblock {\em Iie Transactions}, 47(10):1053--1071, 2015.

\bibitem{cho2014}
Kyunghyun Cho, Bart van Merrienboer, Caglar Gulcehre, Dzmitry Bahdanau, Fethi
  Bougares, Holger Schwenk, and Yoshua Bengio.
\newblock Learning phrase representations using {RNN} encoder--decoder for
  statistical machine translation.
\newblock In {\em Proceedings of the 2014 Conference on Empirical Methods in
  Natural Language Processing (EMNLP)}, pages 1724--1734, 2014.

\bibitem{cooke2006}
SF~Cooke and TVP Bliss.
\newblock Plasticity in the human central nervous system.
\newblock {\em Brain}, 129(7):1659--1673, 2006.

\bibitem{debanne2011}
Dominique Debanne, Emilie Campanac, Andrzej Bialowas, Edmond Carlier, and
  Gis{\`e}le Alcaraz.
\newblock Axon physiology.
\newblock {\em Physiological Reviews}, 91(2):555--602, 2011.

\bibitem{fitzsimonds1997}
Reiko~Maki Fitzsimonds, Hong-jun Song, and Mu-ming Poo.
\newblock Propagation of activity-dependent synaptic depression in simple
  neural networks.
\newblock {\em Nature}, 388(6641):439, 1997.

\bibitem{gers2000}
Felix~A Gers and J{\"u}rgen Schmidhuber.
\newblock Recurrent nets that time and count.
\newblock In {\em Proceedings of the 2000 International Joint Conference on
  Neural Networks (IJCNN). Neural Computing: New Challenges and Perspectives
  for the New Millennium}, volume~3, pages 189--194, 2000.

\bibitem{gerstner2002}
Wulfram Gerstner and Werner~M Kistler.
\newblock {\em Spiking Neuron Models: Single Neurons, Populations, Plasticity}.
\newblock Cambridge University Press, 2002.

\bibitem{guzman2017}
Sandra~M Guzman, Joel~O Paz, and Mary Love~M Tagert.
\newblock The use of {NARX} neural networks to forecast daily groundwater
  levels.
\newblock {\em Water Resources Management}, 31(5):1591--1603, 2017.

\bibitem{hochreiter1997}
Sepp Hochreiter and J{\"u}rgen Schmidhuber.
\newblock Long short-term memory.
\newblock {\em Neural Computation}, 9(8):1735--1780, 1997.

\bibitem{lai2018}
Guo-Kun Lai, Wei-Cheng Chang, Yi-Ming Yang, and Han-Xiao Liu.
\newblock Modeling long-and short-term temporal patterns with deep neural
  networks.
\newblock In {\em Proceedings of the 41st International Conference on Research
  \& Development in Information Retrieval (SIGIR)}, pages 95--104, 2018.

\bibitem{lodish2008}
Harvey Lodish, Arnold Berk, Chris~A Kaiser, Monty Krieger, Matthew~P Scott,
  Anthony Bretscher, Hidde Ploegh, Paul Matsudaira, et~al.
\newblock {\em Molecular Cell Biology}.
\newblock Macmillan, 2008.

\bibitem{mattson1988}
Mark~P Mattson.
\newblock Neurotransmitters in the regulation of neuronal cytoarchitecture.
\newblock {\em Brain Research Reviews}, 13(2):179--212, 1988.

\bibitem{mcculloch1943}
Warren~S McCulloch and Walter Pitts.
\newblock A logical calculus of the ideas immanent in nervous activity.
\newblock {\em The Bulletin of Mathematical Biophysics}, 5(4):115--133, 1943.

\bibitem{niu2012}
Xiao-Xiao Niu and Ching~Y Suen.
\newblock A novel hybrid {CNN-SVM} classifier for recognizing handwritten
  digits.
\newblock {\em Pattern Recognition}, 45(4):1318--1325, 2012.

\bibitem{parcollet2019}
Titouan Parcollet, Mirco Ravanelli, Mohamed Morchid, Georges Linar{\`e}s,
  Chiheb Trabelsi, Renato De~Mori, and Yoshua Bengio.
\newblock Quaternion recurrent neural networks.
\newblock In {\em Proceedings of the 7th International Conference on Learning
  Representations (ICLR)}, 2019.

\bibitem{park2013}
Hyungju Park and Mu-ming Poo.
\newblock Neurotrophin regulation of neural circuit development and function.
\newblock {\em Nature Reviews Neuroscience}, 14(1):7--23, 2013.

\bibitem{pillow2005}
Jonathan~W Pillow, Liam Paninski, Valerie~J Uzzell, Eero~P Simoncelli, and
  EJ~Chichilnisky.
\newblock Prediction and decoding of retinal ganglion cell responses with a
  probabilistic spiking model.
\newblock {\em Journal of Neuroscience}, 25(47):11003--11013, 2005.

\bibitem{schinder2000}
Alejandro~F Schinder and Mu-ming Poo.
\newblock The neurotrophin hypothesis for synaptic plasticity.
\newblock {\em Trends in Neurosciences}, 23(12):639--645, 2000.

\bibitem{shrestha2018}
Sumit~Bam Shrestha and Garrick Orchard.
\newblock Slayer: Spike layer error reassignment in time.
\newblock In {\em Advances in Neural Information Processing Systems 31 (NIPS)},
  pages 1412--1421, 2018.

\bibitem{trabelsi2018}
Chiheb Trabelsi, Olexa Bilaniuk, Ying Zhang, Dmitriy Serdyuk, Sandeep
  Subramanian, Joao~Felipe Santos, Soroush Mehri, Negar Rostamzadeh, Yoshua
  Bengio, and Christopher~J Pal.
\newblock Deep complex networks.
\newblock In {\em Proceedings of the 6th International Conference on Learning
  Representations (ICLR)}, 2018.

\bibitem{vanrullen2005}
Rufin VanRullen, Rudy Guyonneau, and Simon~J Thorpe.
\newblock Spike times make sense.
\newblock {\em Trends in Neurosciences}, 28(1):1--4, 2005.

\bibitem{won2013}
Chee~Sun Won and Robert~M Gray.
\newblock {\em Stochastic Image Processing}.
\newblock Springer Science \& Business Media, 2013.

\bibitem{yang2011}
Hui Yang, Satish~TS Bukkapatnam, and Leandro~G Barajas.
\newblock Local recurrence based performance prediction and prognostics in the
  nonlinear and nonstationary systems.
\newblock {\em Pattern Recognition}, 44(8):1834--1840, 2011.

\end{thebibliography}
\bibliographystyle{plain}


\onecolumn
\appendix
\begin{center}
{\Large\textbf{Supplementary Materials of Flexible Transmitter Network (Appendix)}}
\end{center}
In this Appendix, we provide some detailed supplementary materials the main text, constructed according to the corresponding sections therein.

\section{Holomorphism, Adjoint Variables, and Complex Chain Rule} \label{app:A}

\subsection{Holomorphism}
A complex-valued function $f$ is called \emph{holomorphism} or \emph{analyticity}, if this function is complex-differentiable in its domain. Formally, for any complex-valued point $z_0 $, the limit value $f'(z_0)$ exists, that is,
\[
f'(z_0) = \lim\limits_{\Delta z \to 0} \frac{f(z_0+\Delta z) - f(z_0)}{\Delta z} .
\]

\subsection{Examples for Holomorphism}
Here we provide two examples to illustrate the holomorphism. The first is the most common linear complex-valued function $L(z) = c \cdot z + b$, where $c = c_1 + c_2 \boldsymbol{i}$, $b = b_1 + b_2 \boldsymbol{i}$ and $c_1,c_2,b_1,b_2 \in\mathbb{R}$. So for $z = z_1 + z_2 \boldsymbol{i}$,  $L(z)$ can also be written as:
\[
L(z) = ( c_1 z_1 - c_2 z_2 + b_1) + (c_1 z_2 + c_2 z_1 + b_2) \boldsymbol{i}.
\]

To prove that the linear complex-valued function $L$ is holomorphic, the following result should hold at every point $z$:
\[
\begin{aligned}
L'(z) &= \lim\limits_{\Delta z \to 0} \frac{L(z+\Delta z) - L(z)}{\Delta z} \\
&= \lim\limits_{\Delta z \to 0} \frac{( c_1\Delta z_1 - c_2\Delta z_2) + (c_1\Delta z_2 + c_2\Delta z_1) \boldsymbol{i}}{\Delta z_1 + \Delta z_2\boldsymbol{i}} \\
&= c_1 + c_2 \boldsymbol{i} \\
&= c.
\end{aligned}
\]

The second example is the quadratic function $Q(z) = z^2$. Similarly, $Q$ is holomorphic, according to
\[
\begin{aligned}
Q'(z) &= \lim\limits_{\Delta z \to 0} \frac{Q(z+\Delta z) - Q(z)}{\Delta z} \\
&= \lim\limits_{\Delta z \to 0} \frac{ (2z_1\Delta z_1 + \Delta z_1^2 - 2z_2\Delta z_2 - \Delta z_2^2) + (z_1\Delta z_2 + z_2\Delta z_1 + \Delta z_1 \Delta z_2) \boldsymbol{i} }{\Delta z_1 + \Delta z_2\boldsymbol{i}} \\
&= 2 ( z_1 + z_2 \boldsymbol{i} ) \\
&= 2z.
\end{aligned}
\]

\subsection{Cauchy-Riemann Condition}
Obviously, the limit language is too impractical for identifying a holomorphic function. Fortunately, there is a pair of Cauchy-Riemann equations, which gives a sufficient condition for a complex-valued function to be holomorphic. In detail, $f(z) = u(z_1,z_2) + v(z_1, z_2) \boldsymbol{i}$ is differentiable, where $u$ and $v$ both are real-valued functions, if $f$ satisfies the following two terms:
\begin{itemize}
\item $\frac{\partial u}{\partial z_1}$, $\frac{\partial u}{\partial z_2}$, $\frac{\partial~ v}{\partial z_1}$, and $\frac{\partial v}{\partial z_2}$ are continuous in real domain;
\item The equations $\frac{\partial u}{\partial z_1} = \frac{\partial v}{\partial z_2}$ and $\frac{\partial u}{\partial z_2} = -\frac{\partial v}{\partial z_1}$ hold.
\end{itemize}

\subsection{Adjoint Variables}
Adhering to this line of the Cauchy-Riemann equations, we can find:
\[
\frac{\partial~ (u,v)}{\partial~ (z_1,z_2)} = 
\begin{bmatrix}
u_{z_1} & u_{z_2} \\
v_{z_1} & v_{z_2}
\end{bmatrix} = \begin{bmatrix}
A & B\\
-B & A
\end{bmatrix} .
\]
It is worth noting that the partial derivative matrix of a holomorphic function is anti-symmetric and full-rank. 
\begin{wrapfigure}{r}{0.4\textwidth}
	\centering
	\includegraphics[width=0.4\textwidth]{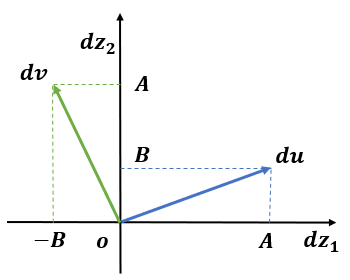}
	\caption{Orthogonality}
	\label{fig:orthogonal}
\end{wrapfigure}
So the following equations hold:
\[
\left\{\begin{aligned}
d~ u(x,y) &= A ~dz_1 + B ~dz_2\\
d~ v(x,y) &= -B ~dz_1 + A ~dz_2
\end{aligned}\right. ,
\]
which means that the gradients / increments of $u,v$ with respect to $z_1,z_2$ are orthogonal, as shown in Figure~\ref{fig:orthogonal}. Furthermore, we can calculate $u$ and $v$ by the integration equations as follows:
\[
\left\{\begin{aligned}
u(z_1,z_2) &= \int_L A ~dz_1 + B ~dz_2 ,\\
v(z_1,z_2) &= \int_L -B ~dz_1 + A ~dz_2 .
\end{aligned}\right.
\]
Obviously, both two integration formulas are path-independent. So we can claim that $u(z_1,z_2)$ is \emph{symplectic symmetric} to $v(z_1,z_2)$. In addition, if we mastered the value or formulation of $u(z_1,z_2)$, we could easily derive $v(z_1,z_2)$. Therefore, we can notice that $r_t = v(z_1,z_2)$ is an adjoint variable of $s_t = u(z_1,z_2)$.

\subsection{A vivid illustration for adjoint variables}
\begin{figure*}[!htb]
\centering
\subfigure[]{
	\begin{minipage}[t]{0.23\textwidth}
		\includegraphics[width=1\textwidth]{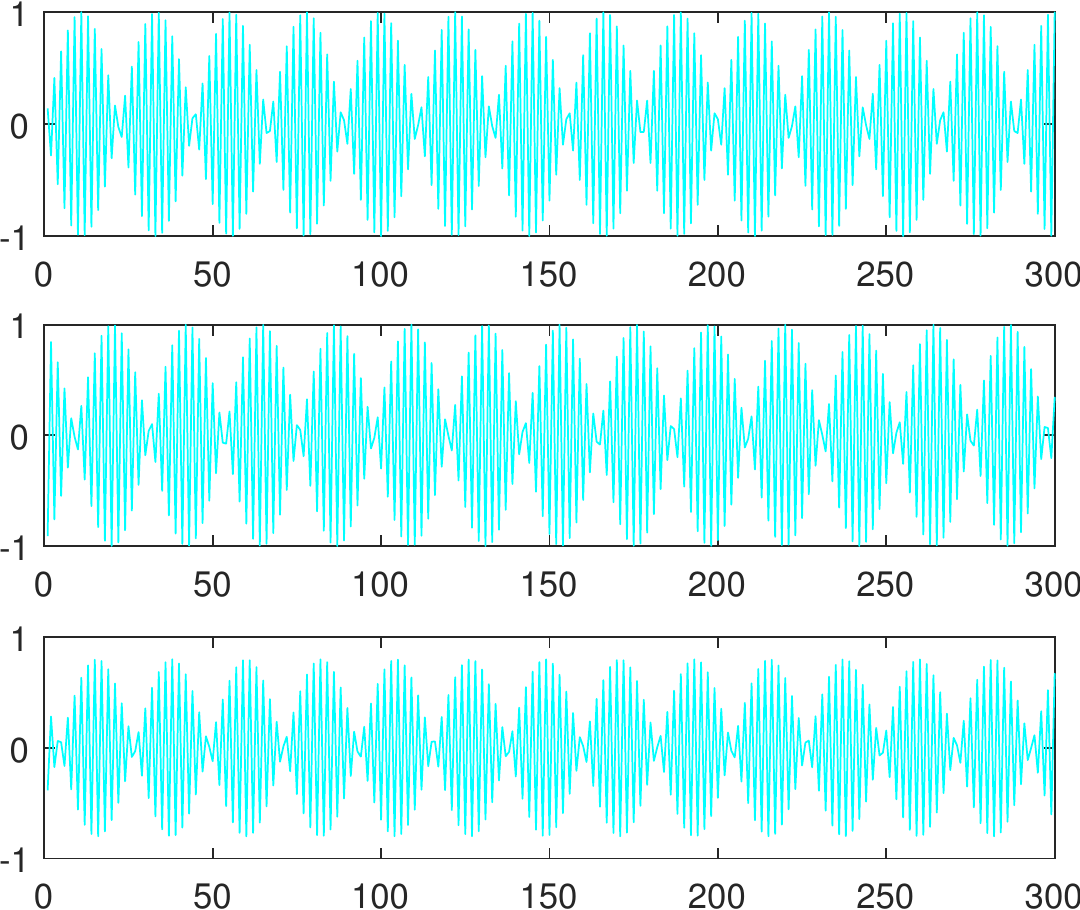}
	\end{minipage}\label{fig:9a}
} \hfill
\subfigure[]{
	\begin{minipage}[t]{0.23\textwidth}
		\includegraphics[width=1\textwidth]{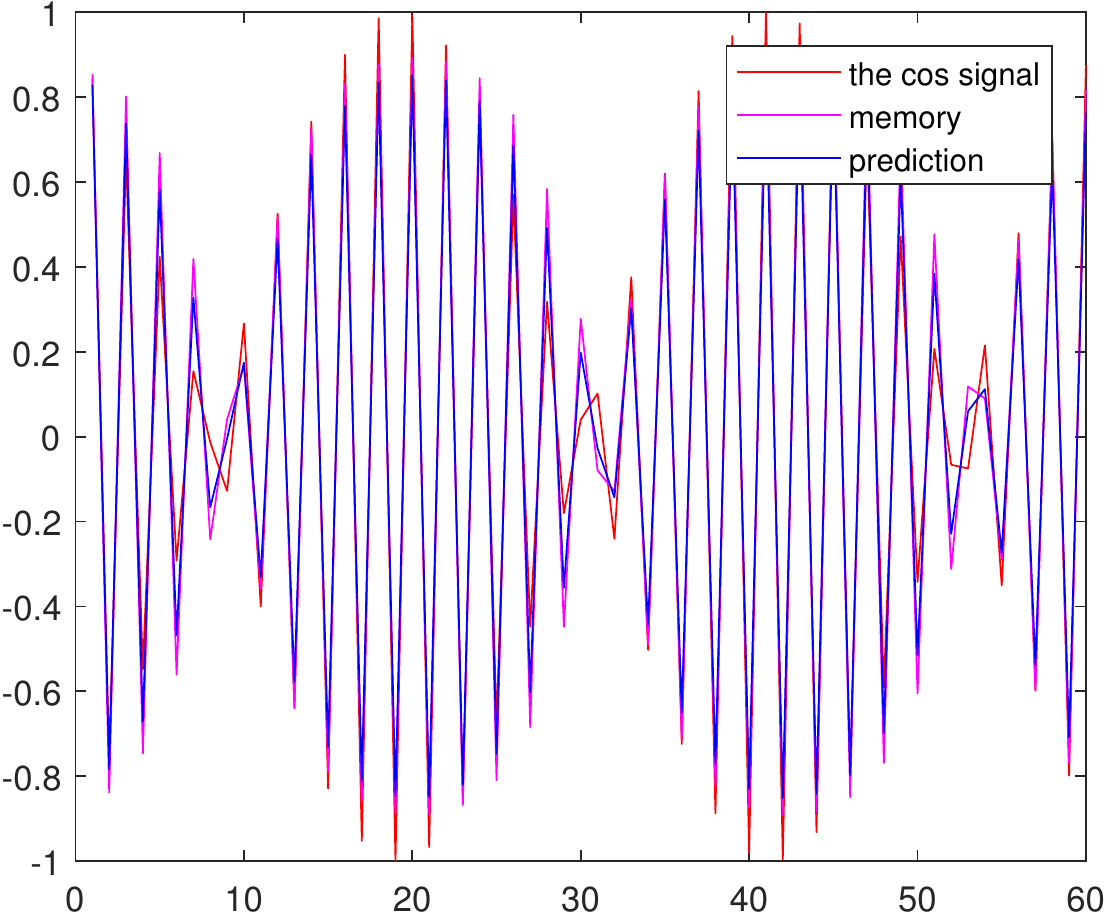}
	\end{minipage}\label{fig:9b}
} \hfill
\subfigure[]{
	\begin{minipage}[t]{0.23\textwidth}
		\includegraphics[width=1\textwidth]{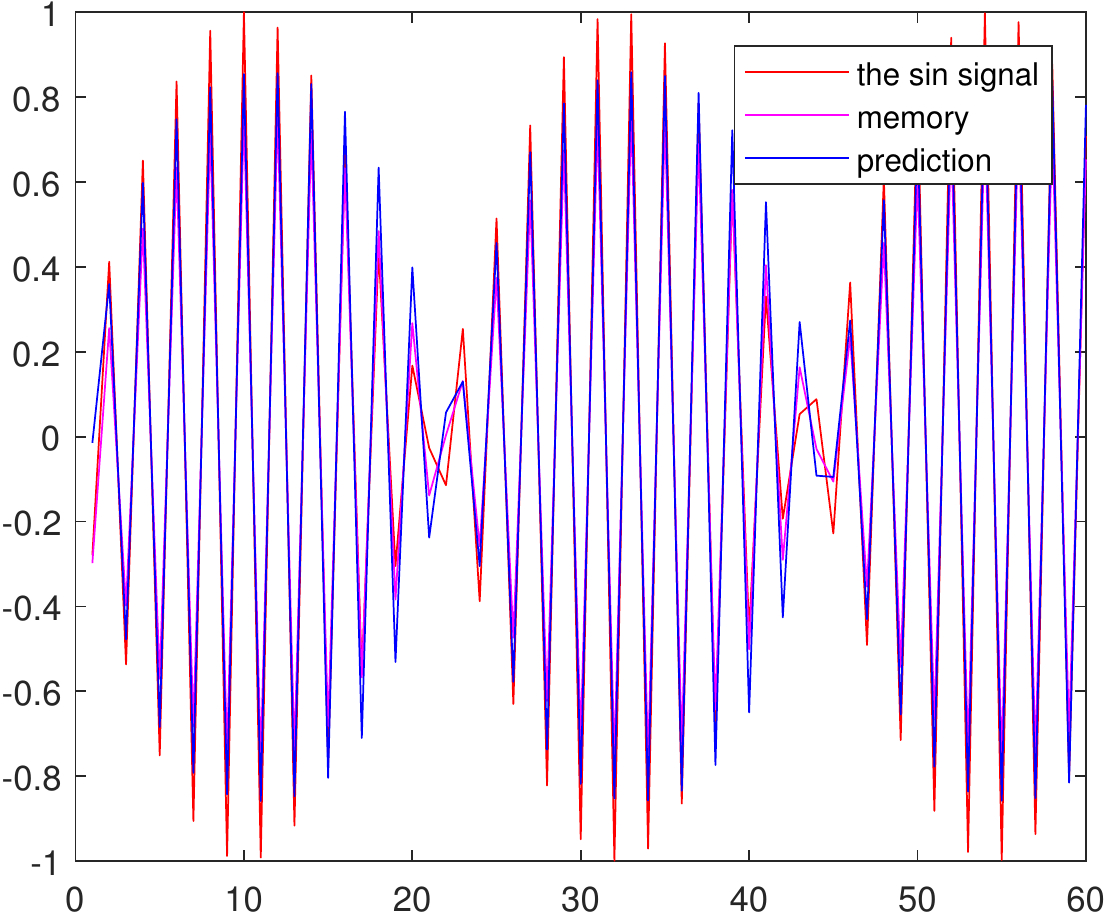}
	\end{minipage}\label{fig:9c}
} \hfill
\subfigure[]{
	\begin{minipage}[t]{0.23\textwidth}
		\includegraphics[width=1\textwidth]{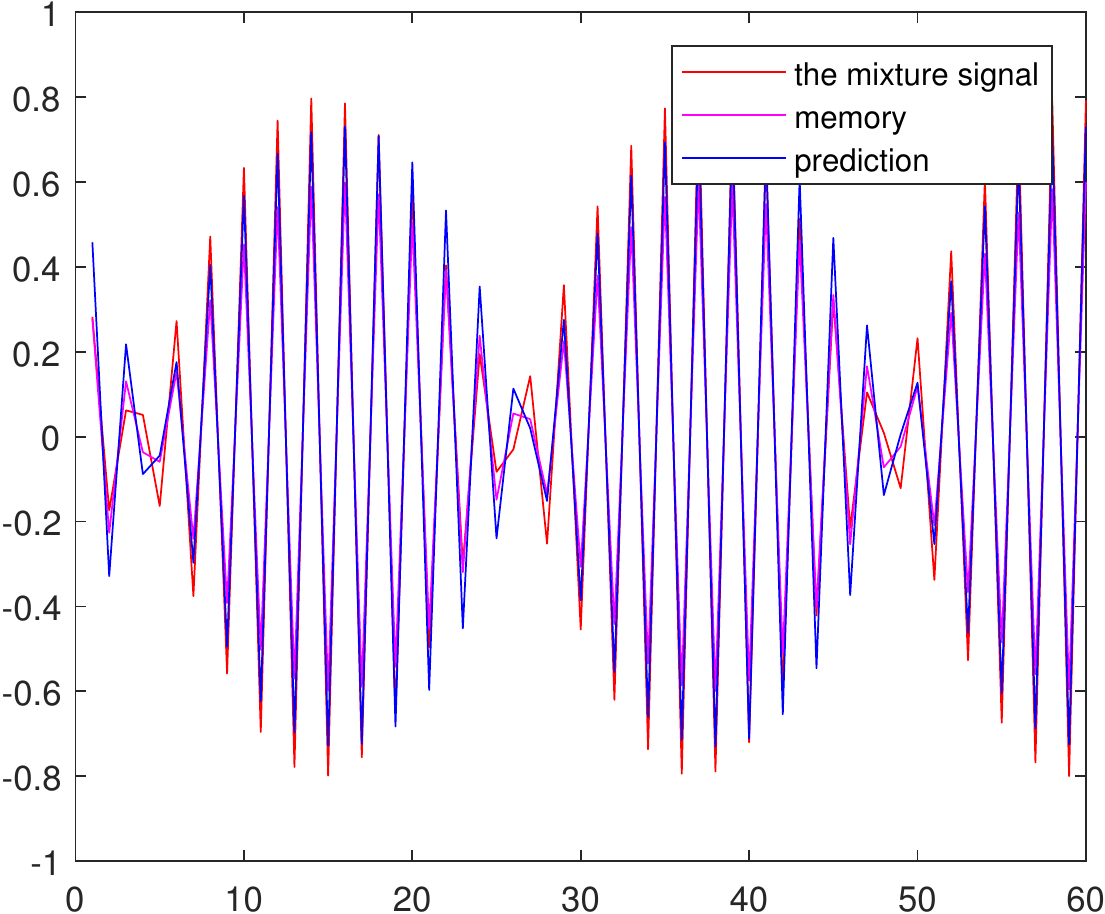}
	\end{minipage}\label{fig:9d}
}
\caption{Figure (a) plots the simulated signals, that is, from top to bottom, the cos function with a period of 3, the sin function with a period of 3, and the mixture functions over 300 timestamps. Figure (b)-(d) display the relation of supervised signals ({\color{red}red}), memory ({\color{magenta}magenta}), and prediction ({\color{blue}blue}), respectively.}
\label{fig:vivid}
\end{figure*}

In order to perform the superiority of the twins' law or the adjoint mechanism in the FT model thoroughly, we design the following experiments. Firstly, we generate 3 signals, that is, a cos function with a period of 3, a sin function with a period of 3, and a mixture of the two aforementioned functions over 300 timestamps, as shown in Figure~\ref{fig:9a}. We employ only one FT neuron to fit these lines. The results are listed in Figure~\ref{fig:9b},~\ref{fig:9c}, and~\ref{fig:9d}, respectively.

Three pairs of parameters are $\left\{\begin{aligned}
\mathbf{W} &= -1.3467 \\
\mathbf{V} &= -0.0943
\end{aligned}\right.$, $\left\{\begin{aligned}
\mathbf{W} &= -1.1420 \\
\mathbf{V} &= -0.1929
\end{aligned}\right.$, and $\left\{\begin{aligned}
\mathbf{W} &= -1.0137 \\
\mathbf{V} &= -0.1988
\end{aligned}\right.$, respectively.

The proposed FT model derives a local recurrent system, thus able to handle spatio-temporal signals. Speaking of this, it is easy to remind us of the typical recurrent neural networks. Different from the RNN's unit formalized by a real-valued function $s_t = g(x_t, s_{t-1};w,v)$, the FT model is dominated by $(s_t,r_t) = f(x_t,r_{t-1};w,v)$. Obviously, the RNN's unit is a special case of the FT model; these two models are equivalent to each other when pre-setting $r_t = s_t$. Therefore, with the same number of parameters $w$ and $v$, the FT model has a broader representational capacity.

\begin{figure}[t]
	\centering
	\subfigure[]{
		\begin{minipage}[t]{0.25\textwidth}
			\includegraphics[width=1\textwidth]{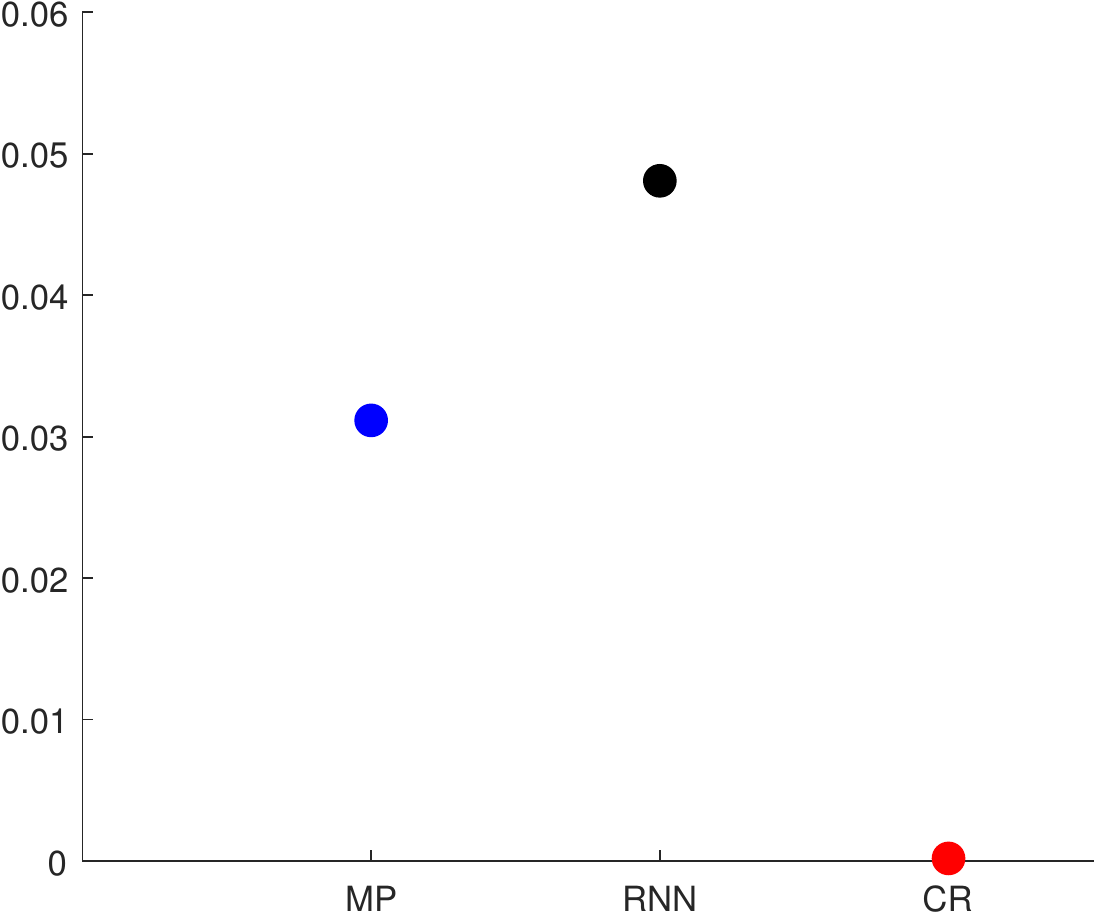}
		\end{minipage}
	}
	\hfill
	\subfigure[]{
		\begin{minipage}[t]{0.25\textwidth}
			\includegraphics[width=1\textwidth]{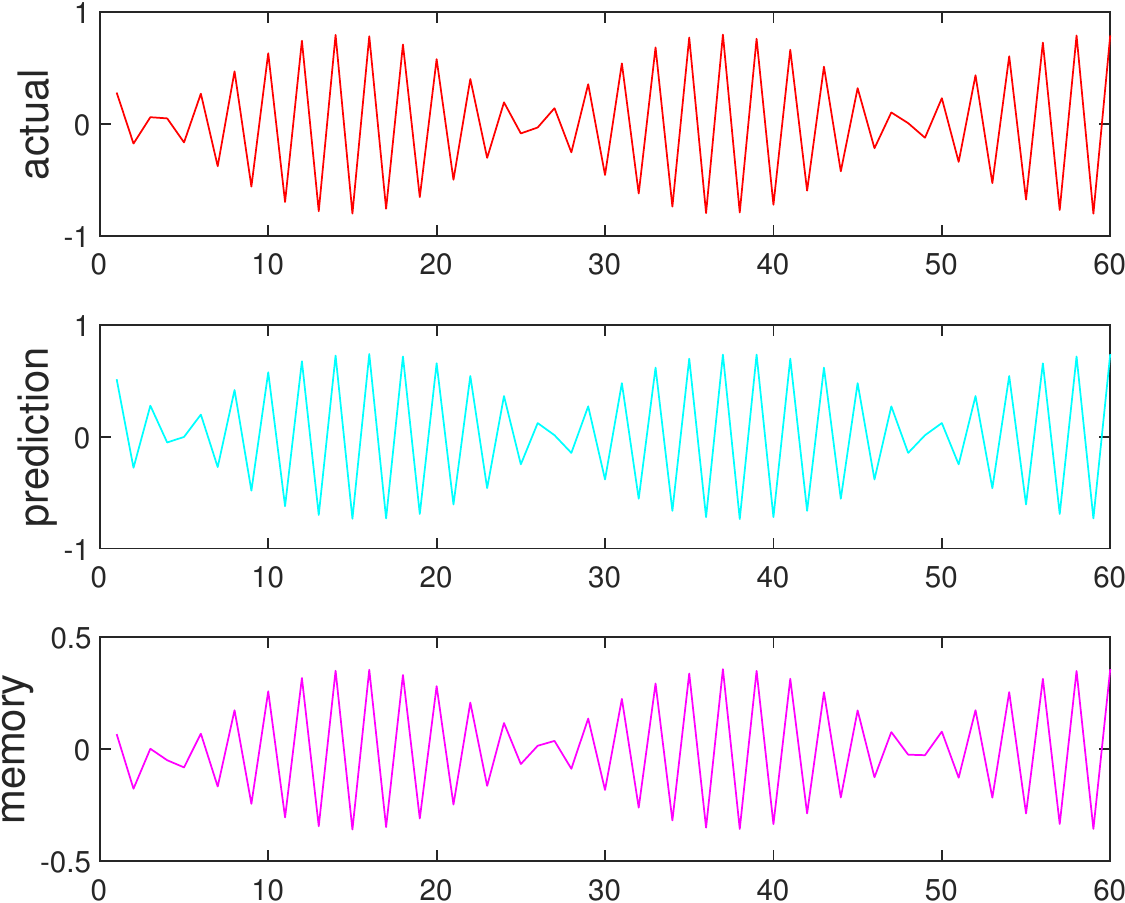}
		\end{minipage}
	}
	\caption{(a) The MSE of the MP model, RNNs' unit, and FT model (CR). (b) The pictures from top to bottom are supervised signals, prediction signals, and memory records of our FT neuron.}
	\label{fig:app_adjoint}
\end{figure}

In order to further demonstrate its superiority thoroughly, we design a simple experiment that takes only one MP model, one RNNs' unit, and one FT neuron (via the complex-valued reaction, i.e., CR), respectively to fit a stimulated curve, and then appraises their performance. The curve is a mixture of a cosine function and a sine function with a period of 3 over 300 timestamps. Here, we use the complex-valued function to implement the FT model, such as Equation~\ref{eq:cr}. The comparative results shown in Figure~\ref{fig:app_adjoint}(a) state that our FT neuron outperforms the other two models, achieving the minimum MSE. It is proved that our FT model can handle more complicated data rather than the conventional MP model and RNNs. 

Next, we boldly speculate about the reasons for the superior performance of the FT model. The idea still revolves around the twins' law or adjoint mechanism. As mentioned above, the real and imaginary parts of a complex-valued function obey the twins' law, that is, $s_t$ and $r_t$ in Equation~\ref{eq:cr} are geminous twins. To illustrate this view, we also investigated the real (prediction) signals and imaginary records of the FT neuron, which are shown in Figure~\ref{fig:app_adjoint}(b). Compared to the prediction signals, the imaginary records evolve with smaller amplitudes but always maintain the correct directions. In other words, the imaginary variable becomes an ``adjoint" one of the real prediction. When FT neuron makes feed-forward forecasting, the ``adjoint" variable would play an auxiliary role and contribute to improving precise performance.

\subsection{Complex Chain Rule}
If $f$ is holomorphic, then
\[
\begin{aligned}
\nabla_z f &= \frac{\partial~ f(z)}{\partial~ z} = \frac{\partial~ u(z_1,z_2) + v(z_1,z_2)\boldsymbol{i}}{\partial~ z_1 + z_2\boldsymbol{i}} \\
&= \frac{\partial~ u}{\partial~ z_1} + \frac{\partial~ u}{\partial z_2 \boldsymbol{i}} + \frac{\partial~ v\boldsymbol{i}}{\partial~ z_1} + \frac{\partial~ v\boldsymbol{i}}{\partial~ z_2\boldsymbol{i}} \\
&= \left( \frac{\partial~ u}{\partial~ z_1} + \frac{\partial~ v}{\partial~ z_2} \right) + \left( \frac{\partial~ v}{\partial~ z_1} - \frac{\partial~ u}{\partial~ z_2} \right)\boldsymbol{i} .
\end{aligned}
\]
If $z_1$ and $z_2$ can be expressed as real-valued holomorphic functions of another complex variable $s = s_1 + s_2 \boldsymbol{i}$, then
\[
\resizebox{1\hsize}{!}{$
\begin{aligned}
\nabla_s f &= \frac{\partial~ f(z)}{\partial~ s} = \frac{\partial~ u(z_1,z_2) + v(z_1,z_2)\boldsymbol{i}}{\partial~ s_1 + s_2\boldsymbol{i}} \\
&= \left( \frac{\partial~ u}{\partial~ s_1} + \frac{\partial~ v}{\partial~ s_2} \right) + \left( \frac{\partial~ v}{\partial~ s_1} - \frac{\partial~ u}{\partial~ s_2} \right)\boldsymbol{i} \\
&= \left( \frac{\partial~ u}{\partial~ z_1} \frac{\partial~ z_1}{\partial~ s_1} + \frac{\partial~ u}{\partial~ z_2} \frac{\partial~ z_2}{\partial~ s_1} + \frac{\partial~ v}{\partial~ z_1} \frac{\partial~ z_1}{\partial~ s_2} + \frac{\partial~ v}{\partial~ z_2} \frac{\partial~ z_2}{\partial~ s_2} \right) + \left( \frac{\partial~ v}{\partial~ z_1} \frac{\partial~ z_1}{\partial~ s_1} + \frac{\partial~ v}{\partial~ z_2} \frac{\partial~ z_2}{\partial~ s_1} - \frac{\partial~ u}{\partial~ z_1} \frac{\partial~ z_1}{\partial~ s_2} - \frac{\partial~ u}{\partial~ z_2} \frac{\partial~ z_2}{\partial~ s_2} \right)\boldsymbol{i} .
\end{aligned}
$}
\]

\section{Activations} \label{app:activations}
Here we are going to introduce some activation functions that can generally be used in FTNet. Thanks to Equation~\ref{eq:activation}, the vectorized representation has been converted into:
\[
\boldsymbol{s}_t + \boldsymbol{r}_t \boldsymbol{i} = \sigma \left(~ ( a \mathbf{W} \boldsymbol{x}_t - b \mathbf{V} \boldsymbol{r}_{t-1} ) + ( b \mathbf{W} \boldsymbol{x}_t + a \mathbf{V} \boldsymbol{r}_{t-1}) \boldsymbol{i} ~\right).
\]
An intuitive idea is to decompose the activation function $\sigma$ into two real-valued nonlinear functions that are activated with respect to the real and imaginary parts, respectively, that is, $\sigma  = \sigma_{real} + \sigma_{imag} \boldsymbol{i}$, where $\sigma_{real}$ and $\sigma_{imag}$ are real-valued activation functions, such as the \emph{sigmoid} and \emph{tanh} functions. So Equation~\ref{eq:activation} becomes:
\[
\boldsymbol{s}_t + \boldsymbol{r}_t \boldsymbol{i} = \sigma_{real}( a \mathbf{W} \boldsymbol{x}_t - b \mathbf{V} \boldsymbol{r}_{t-1} ) + \sigma_{imag}( b \mathbf{W} \boldsymbol{x}_t + a \mathbf{V} \boldsymbol{r}_{t-1}) ) \boldsymbol{i}.
\]

\begin{figure}[t]
	\centering
	\subfigure[]{
		\begin{minipage}[t]{0.23\textwidth}
			\includegraphics[width=1\textwidth]{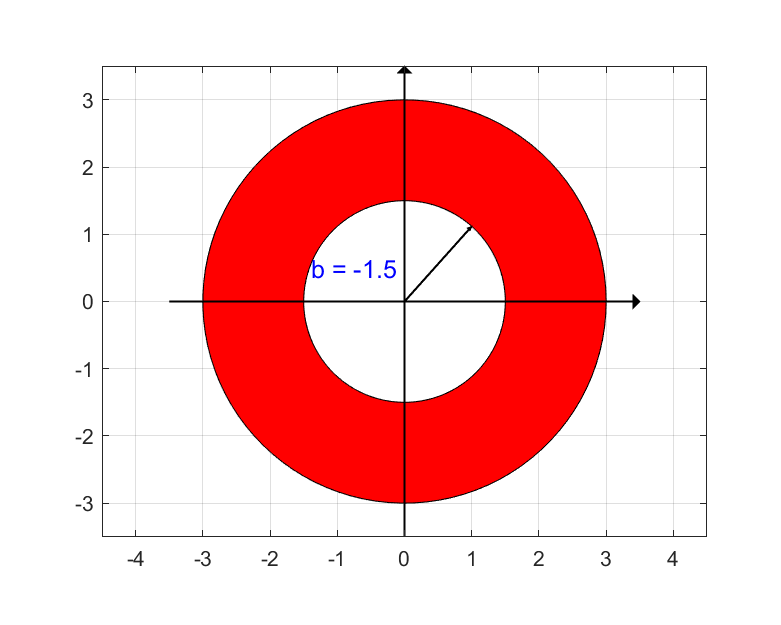}
		\end{minipage}
	}
	\hfill
	\subfigure[]{
		\begin{minipage}[t]{0.23\textwidth}
			\includegraphics[width=1\textwidth]{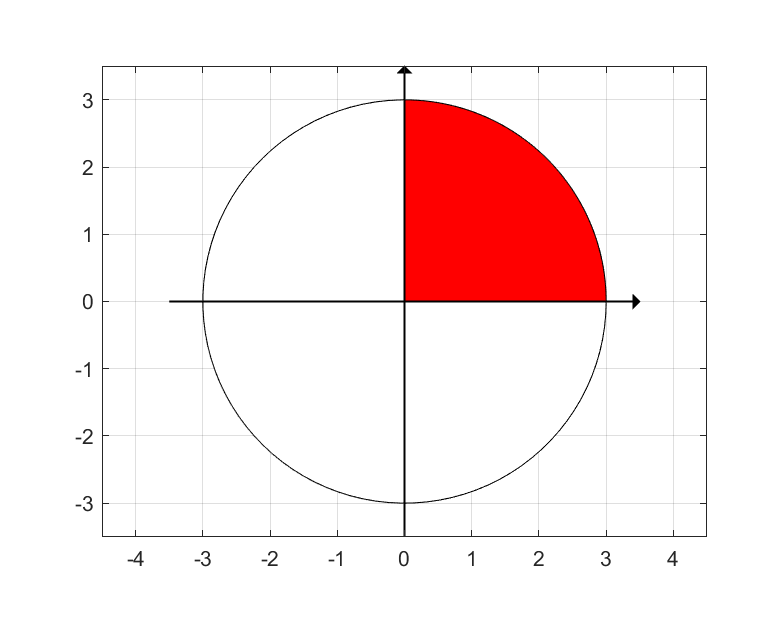}
		\end{minipage}
	}
	\hfill
	\subfigure[]{
		\begin{minipage}[t]{0.23\textwidth}
			\includegraphics[width=1\textwidth]{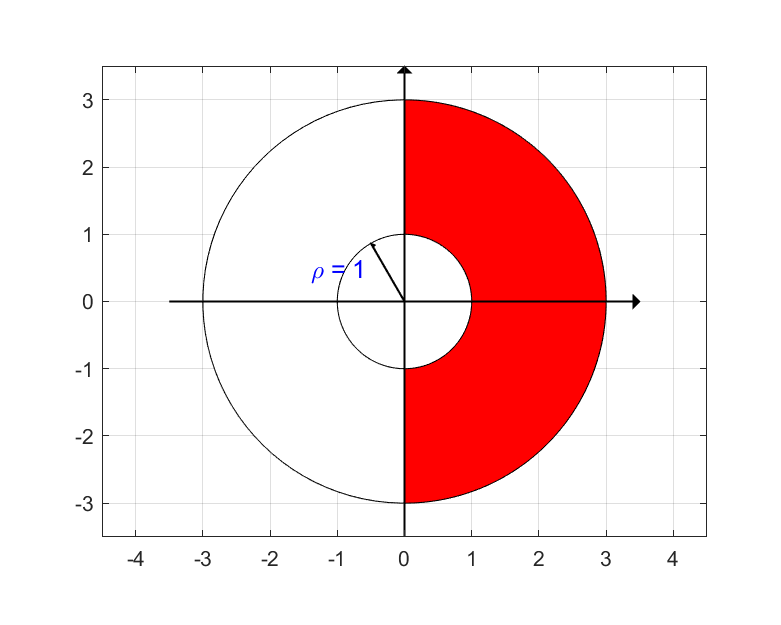}
		\end{minipage}
	}
	\hfill
	\subfigure[]{
		\begin{minipage}[t]{0.23\textwidth}
			\includegraphics[width=1\textwidth]{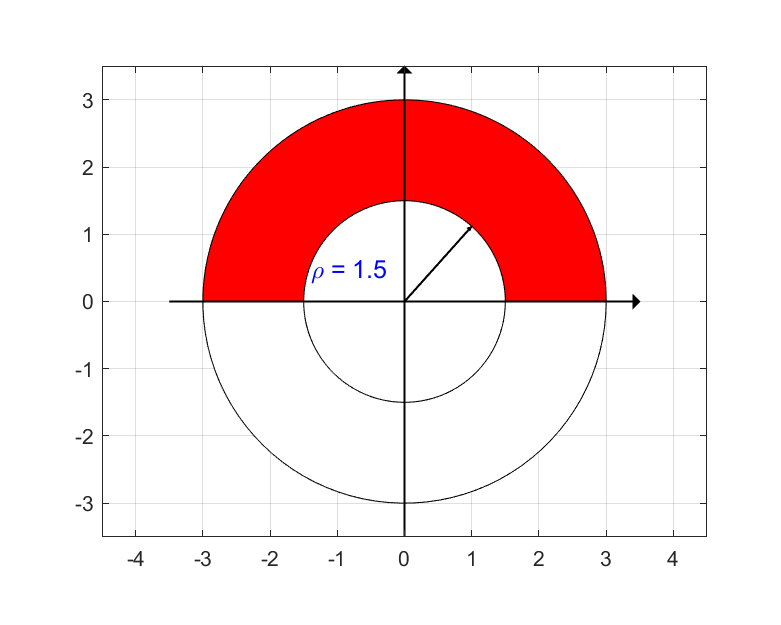}
		\end{minipage}
	}
	\caption{The illustrations of ReLu activations: (a) modReLU with $b=-1.5$; (b) zReLU; (c) PReLU with $\rho=1$, $\theta^{*}=[-\frac{\pi}{2},\frac{\pi}{2}]$; (d) PReLU with $\rho=\frac{3}{2}$, $\theta^{*}=[0,\pi]$.}
	\label{fig:activations}
\end{figure}
Apart from this, FTNet also allow non-trivial activations in the complex domain. Indeed, efforts have been made in complex-valued activations, such as the modReLU~\cite{arjovsky2016} and zReLU~\cite{trabelsi2018}. 
\[
modReLU(z) = ReLU(|z|+b)e^{(\theta_z \boldsymbol{i})} = \left\{ \begin{aligned}
( |z| + b )\frac{z}{|z|},~~ & \text{if}~~ |z|+b \geq0 ~, \\
0~~~~~~~~,~~ & \text{otherwise} ~,
\end{aligned}\right.
\]
where $z \in \mathbb{C}$, $\theta_z$ is the phase of $z$, and $b \in \mathbb{R}$ is a learnable parameter.
\[
zReLU(z) = \left\{ \begin{aligned}
z,~~ & \text{if}~~ \theta_z \in [0,\pi/2] ~, \\
0,~~ & \text{otherwise} ~.
\end{aligned}\right.
\]
Here, we propose an alternative complex-valued ReLU activation, the Polar ReLU (PReLU). PReLU is a point-wise nonlinear function, which limits the radius and angle of a complex number in the polar coordinate system, defined as:
\[
PReLU(z) = \max\{|z|,\rho\} \cdot e^{\mathbb{I}(\theta_z \in \theta^{*}) \boldsymbol{i}} = \left\{ \begin{aligned}
z,~~ & \text{if}~~ |z|\geq\rho ~\text{and}~ \theta_z\in ~\theta^{*}, \\
0,~~ & \text{otherwise} ~.
\end{aligned}\right.
\]
$\rho$ is a learnable or predefined parameter, indicating the minimum radius of the complex value $z$, $\theta_{\boldsymbol{z}}$ is the phase of $\boldsymbol{z}$, and $\theta^{*}$ restricts the allowed excitation angle.

Figure~\ref{fig:activations} illustrates the working mechanisms of these aforementioned complex-valued activations from the perspective of geometric manifolds. The modReLU function creates a ``dead circle" with a learnable radius $b$, while zReLU emphasizes the allowed excitation phase of $z$. Our proposed PReLU is able to juggle both angle and radius. As $|\boldsymbol{z}|$ is always greater than $\rho$, PReLU could create a ``dead circle" with radius $\rho$ and centre $(0,0)$. Additionally, PReLU also restricts the allowed excitation angel by a pre-given $\theta^{*}$. Thus, PReLU has greater flexibility and potential, and becomes an awesome indicator to evaluate the importance of radius and phase in complex-valued activations. 

\subsection{Comparsion Experiments for Activations} \label{app:simluated}
\begin{figure*}[t]
	\centering
	\subfigure[]{
		\begin{minipage}[t]{0.23\textwidth}
			\includegraphics[width=1\textwidth]{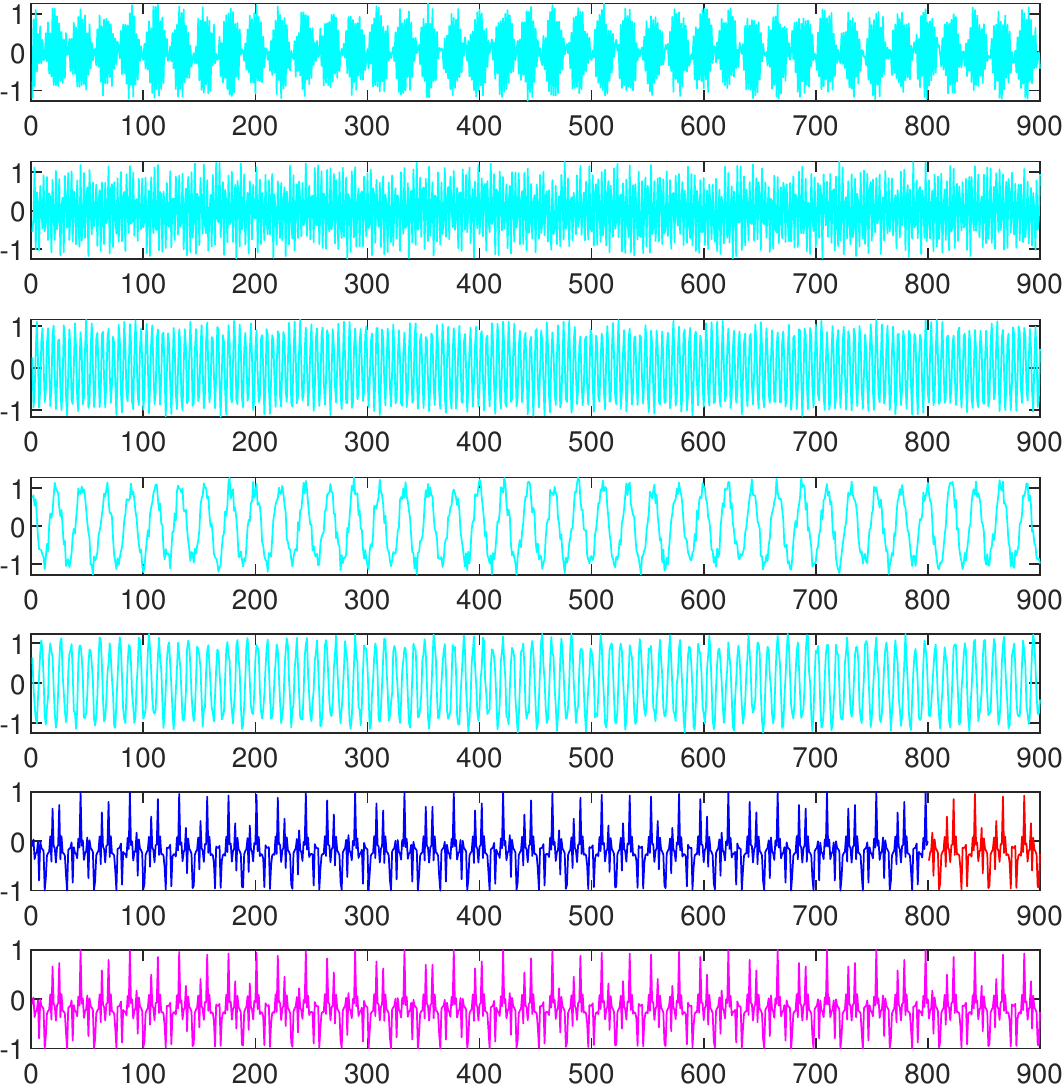}
		\end{minipage}\label{fig:simulation_a}
	} \hfill
	\subfigure[]{
		\begin{minipage}[t]{0.23\textwidth}
			\includegraphics[width=1\textwidth]{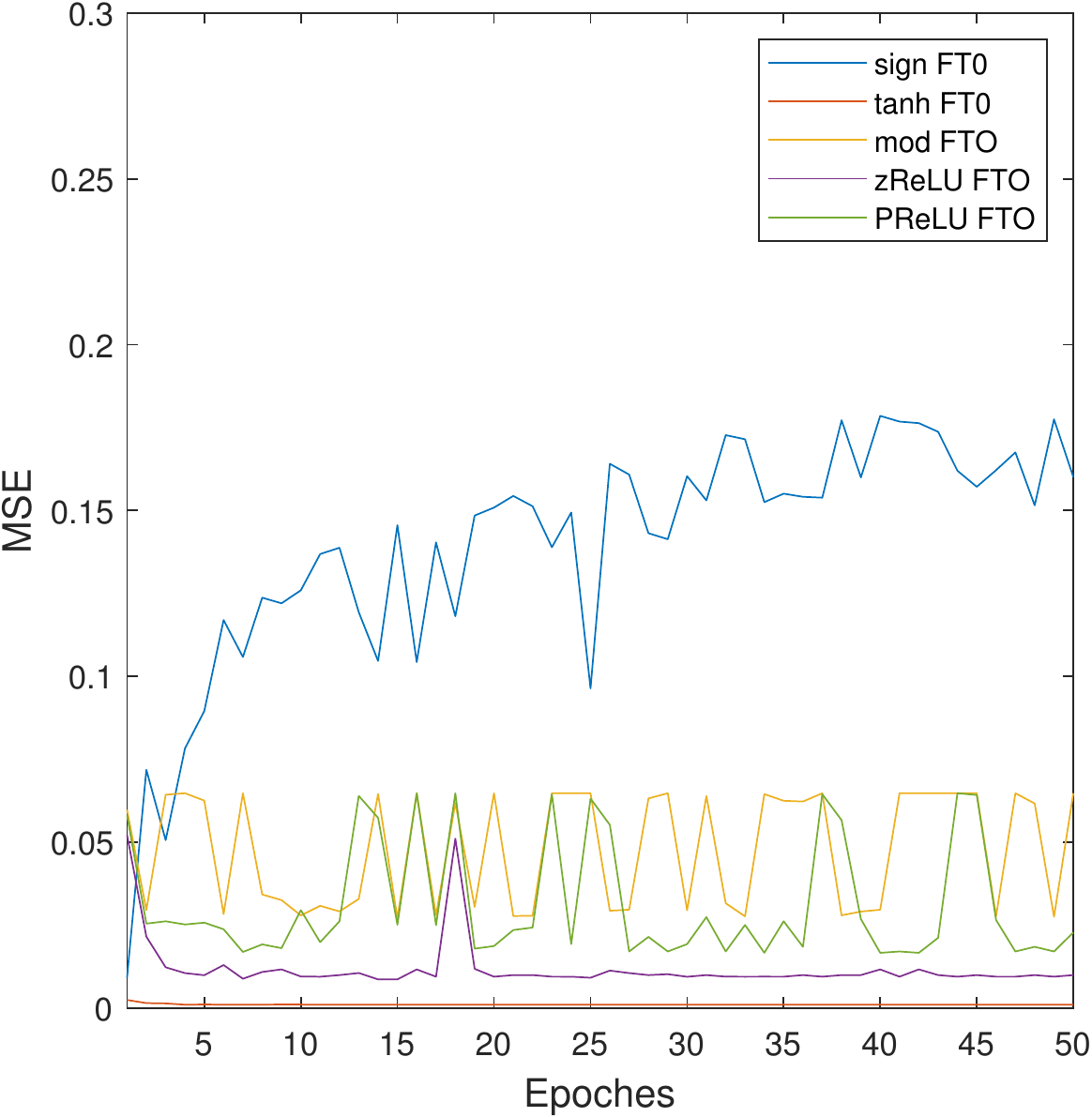}
		\end{minipage}\label{fig:simulation_b}
	} \hfill
	\subfigure[]{
		\begin{minipage}[t]{0.23\textwidth}
			\includegraphics[width=1\textwidth]{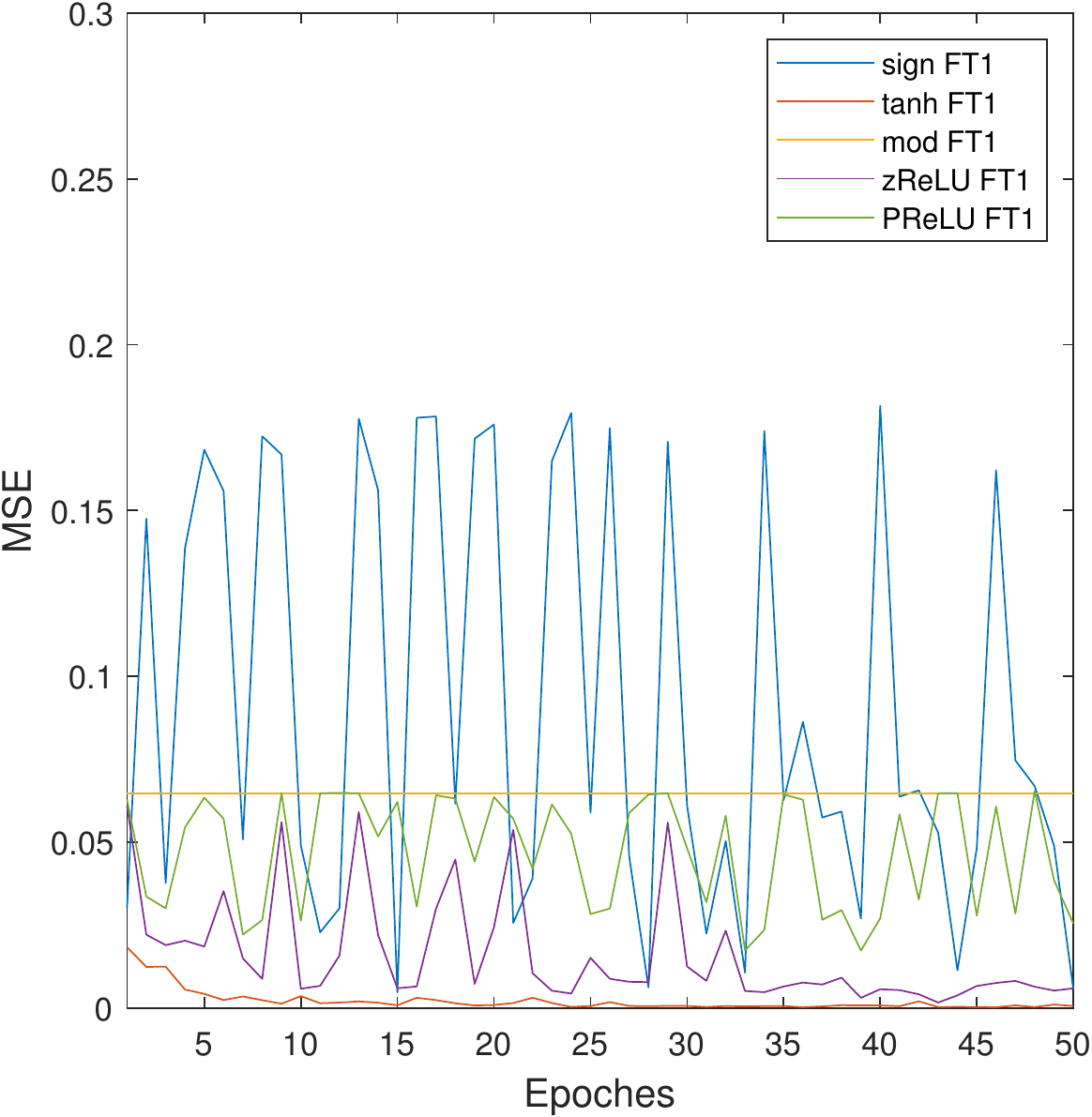}
		\end{minipage}\label{fig:simulation_c}
	} \hfill
	\subfigure[]{
		\begin{minipage}[t]{0.23\textwidth}
			\includegraphics[width=1\textwidth]{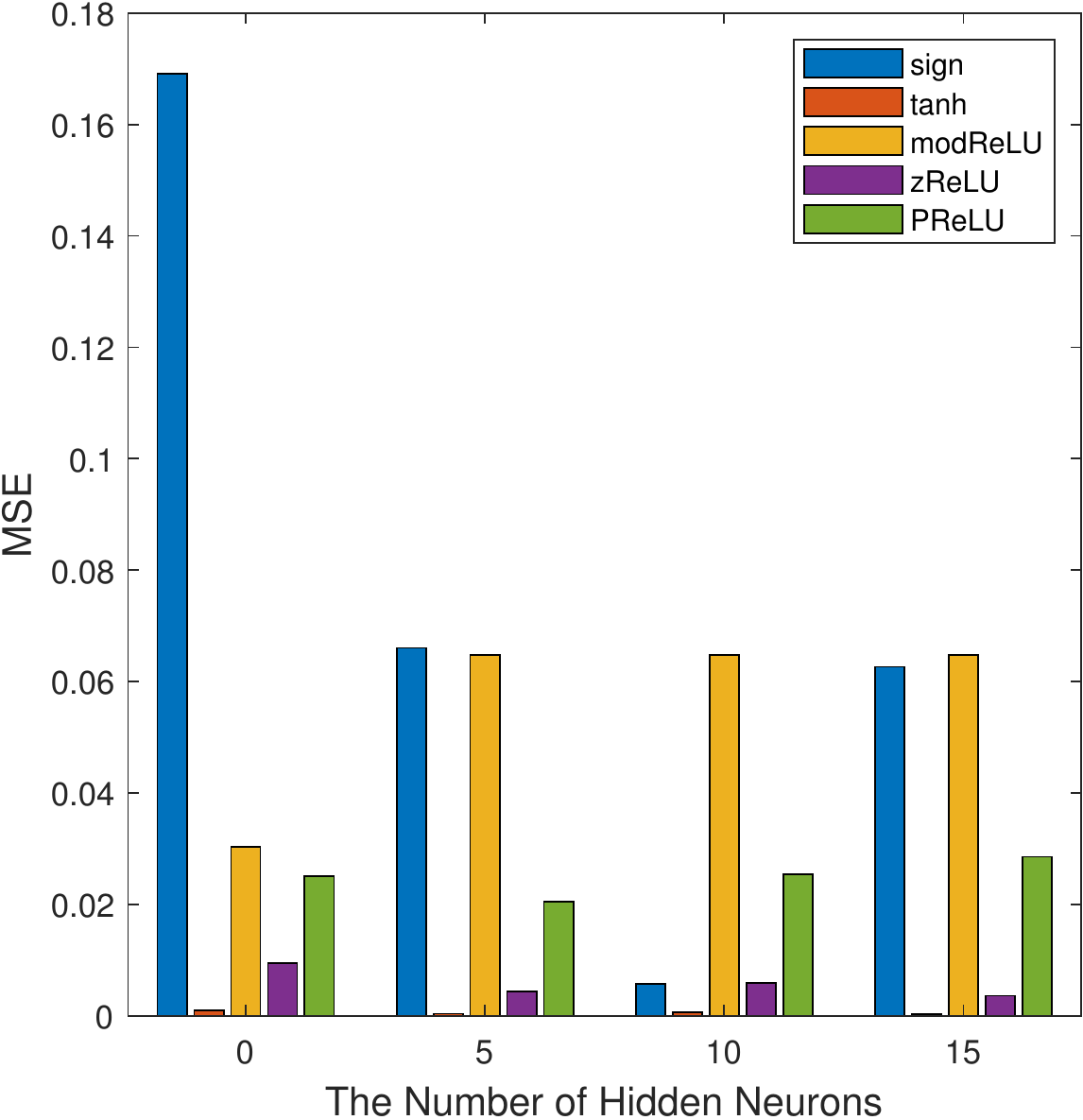}
		\end{minipage}\label{fig:simulation_d}
	}
	\caption{The subplots of Figure (a) from top to bottom are five component signals, mixing stimulated signals that are divided for training({\color{blue}blue}) and testing({\color{red}red}), and the supervised signals. Figure (b) shows the testing MSE of FT0 with a variety of activations ($sigmoid$, $tanh$, $modReLU$, $zReLU$, and $PReLU$) evolves as training iteration increases. Figure (c) shows the comparative results about FT1. Figure (d) displays the effect of neuron quantity on the performance of our FTNet.}
	\label{fig:simulation}
\end{figure*}

We first explore the performance of our FTNet with different configurations on simulated data. The simulation data is generated by aggregating five cosine functions (each function with a period of 3-7) over 900 timestamps. For practice, every component cosine function was fused with a noise signal uniformly sampled with 15\% - 30\% amplitude, which are illustrated in Figure~\ref{fig:simulation_a}. The supervised signals are the mixture of these cosine functions without noise. And we trained FTNet with the first 800 points ({\color{blue}blue}) and forecast the future 100 points ({\color{red}red}).

We ran the FT0 net with five activation functions ($sigmoid$, $tanh$, $modReLU$, $zReLU$, and $PReLU$) until convergence on the test data. Among that, the radius and phase of these complex-valued activations are preset to 0.3 and [0, $\pi/2$], respectively. The experimental results that evolve as training iteration increases are plotted in Figure~\ref{fig:simulation_b}. All the FTNets  except the $sigmoid$ function perform laudably well; however, $tanh$ performs better, achieving the best MSE. Besides, $modReLU$ has unstable performance, while $zReLU$ performs better than $PReLU$. Considering the influence of the cascade structure, we copy this experiment on FT1 with 10 hidden neurons, and display experimental results in Figure~\ref{fig:simulation_c}. The $tanh$ function also outperforms other activations, the second best is $zReLU$, and $PReLU$ is close behind. The remaining two activations perform very poorly; FTNet with the $sigmoid$ activation seems to be strenuous to convergence, while $zReLU$ is even invalid. Besides, we also investigate the effect of the number of hidden neurons. The results are shown in Figure~\ref{fig:simulation_d}. There is a split effect of neuron quantity on the performance of our FTNet when using different activation functions; most models perform better in the case of 10 hidden neurons, whereas $modReLU$ does without the hidden layer and the change in neuron quantity has little effect on its performance. In addition, both $tanh$ and $zReLU$ are able to convincingly overtake the baseline 0.02. $PReLU$ achieves a lower MSE, but there is still a certain gap compared with the former two. The networks with $sigmoid$ and $modReLU$ activations have the poor performance.

In summary, we can conclude that resorting $tanh$ as the activation function is able to attain the best performance, whether using FT0 or FT1 architecture. By contrast, the $sigmoid$ and $modReLU$ functions perform unfavorably. The performance of $zReLU$ is a little better than that of $PReLU$. We conjecture the reason is that the radius may be more susceptible and important than the phase for complex-valued activations. So for the FTNet configurations in the real-world tasks, we employ $tanh$ for activation function and 0.01 for learning rate. Other hyper-parameters cannot be fixed across tasks, otherwise, the performance will be embarrassingly unsatisfactory. Therefore, we examine a variety of configurations on the testing data set and pick out the one with the best performance.

\section{The CBP Algorithm} \label{app:CBP}
In this work, we present the CBP algorithm for training the FTNet, which is an extension of the common back-propagation in the complex domain. The core idea of CBP is to take the neurotrophin density as an implicit variable, so that the desired gradients become a function of the partial derivative of $r_t$ with respect to the connection parameters $\mathbf{W}$ and $\mathbf{V}$. Before that, we first review the feed-forward procedure of FTNet:
\[
\left\{\begin{aligned}
& \boldsymbol{s}^0_t = \boldsymbol{x}_t, \\
& \boldsymbol{s}^l_t + \boldsymbol{r}^l_t \boldsymbol{i} = \sigma ( \boldsymbol{\alpha}^l_t + \boldsymbol{\beta}\boldsymbol{i}^l_t ), \\
& \boldsymbol{y}_t = \boldsymbol{s}_t^L, \\
\end{aligned}\right.
\]
where 
\[
\left\{\begin{aligned}
\boldsymbol{\alpha}^l_t(i) &= a \left( \sum_{k=1}^{n_{l-1}} \mathbf{W}^l(i,k) \boldsymbol{s}^{l-1}_t(k) \right)  - b \left( \sum_{h=1}^{n_l} \mathbf{V}^l(i,h) \boldsymbol{r}_{t-1}(h) \right), \\
\boldsymbol{\beta}^l_t(i) &= b  \left( \sum_{k=1}^{n_{l-1}} \mathbf{W}^l(i,k) \boldsymbol{s}^{l-1}_t(k) \right) + a \left( \sum_{h=1}^{n_l} \mathbf{V}^l(i,h) \boldsymbol{r}_{t-1}(h) \right). \\
\end{aligned}\right.
\]
So for the case that $i=j$, it's easy to obtain:
\[
\left\{\begin{aligned}
\frac{\partial~ \boldsymbol{\alpha}^l_t(i)}{\partial~ \mathbf{W}^l(i,k)} &= a \boldsymbol{s}_t^{l-1}(k) - b \left( \sum_{h=1}^{n_l} \mathbf{V}^l(i,h) ~\frac{\partial~ \boldsymbol{r}_{t-1}^l(h)}{\partial~ \mathbf{W}^l(i,k)} \right), \\
\frac{\partial~ \boldsymbol{\alpha}^l_t(i)}{\partial~ \mathbf{V}^l(i,h)} &= -b \left( \boldsymbol{r}_{t-1}^l(h) + \sum_{h=1}^{n_l} \mathbf{V}^l(i,h) ~\frac{\partial~ \boldsymbol{r}_{t-1}^l(h)}{\partial~ \mathbf{V}^l(i,h)} \right), \\
\end{aligned}\right.
\]
and
\[
\left\{\begin{aligned}
\frac{\partial~ \boldsymbol{\beta}^l_t(i)}{\partial~ \mathbf{W}^l(i,k)} &= b \boldsymbol{s}_t^{l-1}(k) + a \left( \sum_{h=1}^{n_l} \mathbf{V}^l(i,h) ~\frac{\partial~ \boldsymbol{r}_{t-1}^l(h)}{\partial~ \mathbf{W}^l(i,k)} \right), \\
\frac{\partial~ \boldsymbol{\beta}^l_t(i)}{\partial~ \mathbf{V}^l(i,h)} &= a \left( \boldsymbol{r}_{t-1}^l(h) + \sum_{h=1}^{n_l} \mathbf{V}^l(i,h) ~\frac{\partial~ \boldsymbol{r}_{t-1}^l(h)}{\partial~ \mathbf{V}^l(i,h)} \right). \\
\end{aligned}\right.
\]
Ideally, both $\boldsymbol{\alpha}_t^l(i)$ and $\boldsymbol{\beta}_t^l(i)$ are influenced by the ``seemingly independent'' concentration parameters $\mathbf{W}^l(j,k)$ and $\mathbf{V}^l{j,h}$. Regarding the imaginary parts (neurotrophin densities) $\boldsymbol{r}_t^l$ as implicit variables, so for the case that $j \neq i$, we have:
\[
\left\{\begin{aligned}
\frac{\partial~ \boldsymbol{\alpha}^l_t(i)}{\partial~ \mathbf{W}^l(j,k)} &= - b \left( \sum_{h=1}^{n_l} \mathbf{V}^l(j,h) ~\frac{\partial~ \boldsymbol{r}_{t-1}^l(h)}{\partial~ \mathbf{W}^l(j,k)} \right), \\
\frac{\partial~ \boldsymbol{\alpha}^l_t(i)}{\partial~ \mathbf{V}^l(j,h)} &= -b \left( \sum_{h=1}^{n_l} \mathbf{V}^l(j,h) ~\frac{\partial~ \boldsymbol{r}_{t-1}^l(h)}{\partial~ \mathbf{V}^l(j,h)} \right), \\
\end{aligned}\right.
\]
and
\[
\left\{\begin{aligned}
\frac{\partial~ \boldsymbol{\beta}^l_t(i)}{\partial~ \mathbf{W}^l(j,k)} &= a \left( \sum_{h=1}^{n_l} \mathbf{V}^l(j,h) ~\frac{\partial~ \boldsymbol{r}_{t-1}^l(h)}{\partial~ \mathbf{W}^l(j,k)} \right), \\
\frac{\partial~ \boldsymbol{\beta}^l_t(i)}{\partial~ \mathbf{V}^l(j,h)} &= a \left( \sum_{h=1}^{n_l} \mathbf{V}^l(j,h) ~\frac{\partial~ \boldsymbol{r}_{t-1}^l(h)}{\partial~ \mathbf{V}^l(j,h)} \right). \\
\end{aligned}\right.
\]

CBP is slightly different to the back-propagation procedure of recurrent networks. The core difference, or equally challenge of FTNet is the errors caused by the concentration parameters $\mathbf{W}^l$ and $\mathbf{V}^l$ are transmitted not only to the real parts (stimuli signals) $\boldsymbol{s}_t^l$ but also to the imaginary parts (receptor strengths) $\boldsymbol{r}_t^l$. So it is difficult to compute the recurrent derivatives $\frac{\partial~ \boldsymbol{s}_t^l}{\partial~ \boldsymbol{s}_{t-1}^l}$ in FTNet. To tackle this challenge, we take the imaginary parts (neurotrophin densities) $\boldsymbol{r}_t^l$ as implicit variables. So by exploiting the partial derivatives $\left( \frac{\partial~ \boldsymbol{\alpha}^l_t}{\partial~ \mathbf{W}^l}, \frac{\partial~ \boldsymbol{\alpha}^l_t}{\partial~ \mathbf{V}^l} \right)$ and $\left( \frac{\partial~ \boldsymbol{\beta}^l_t}{\partial~ \mathbf{W}^l}, \frac{\partial~ \boldsymbol{\beta}^l_t}{\partial~ \mathbf{V}^l} \right)$, we can collect all the errors caused by $\mathbf{W}^l$ and $\mathbf{V}^l$. Besides, CBP formulates the relation between the gradients $\left( \mathbf{\nabla_{W^l} E}_t, \mathbf{\nabla_{V^l} E}_t \right)$ and the initialization errors of neurotrophin densities $\boldsymbol{r}_0^l$. It is worth to note that before prediction (in the time interval $[0,T]$), we still need to update the imaginary parts (neurotrophin densities $\boldsymbol{r}_{0}^l, \cdots, \boldsymbol{r}_{T}^l$ ) according to:
\[
\left\{\begin{aligned}
& \boldsymbol{s}^0_t = \boldsymbol{x}_t, \\
& \boldsymbol{r}_t^l = \sigma\left( b \hat{\mathbf{W}^l} \boldsymbol{s}_t^{l-1} + a \hat{\mathbf{V}^l} \boldsymbol{r}_{t-1}^l \right), \\
\end{aligned}\right.
\]
and reset the imaginary errors $\left( \frac{\partial~ \boldsymbol{r}^l_T}{\partial~ \mathbf{W}^l}, \frac{\partial~ \boldsymbol{r}^l_T}{\partial~ \mathbf{V}^l} \right)$ as zeros.

The computation complexity of CBP is large, since for each $l$ and $t$, we should at least calculate four tensors, that is, $\frac{\partial~ \boldsymbol{\alpha}^l_t}{\partial~ \mathbf{W}^l}, \frac{\partial~ \boldsymbol{\alpha}^l_t}{\partial~ \mathbf{V}^l}, \frac{\partial~ \boldsymbol{\beta}^l_t}{\partial~ \mathbf{W}^l},$ and $\frac{\partial~ \boldsymbol{\beta}^l_t}{\partial~ \mathbf{V}^l}$. In practice, we suggest to omits the gradients of the case $j \neq i$ for calculation convenience.

Finally,  we have to admit that FTNet also has gradient explosion and vanishing, which is also something we will strive to improve in the future.

\section{Date Sets and Configurations} \label{app:experiments}
The simulation data in subsection~\ref{app:simluated} is generated by aggregating five cosine functions (each function with a period of 3-7) over 900 timestamps. For practice, every component cosine function was fused with a noise signal uniformly sampled with 15\% - 30\% amplitude, which are illustrated in Figure~\ref{fig:simulation_a}. The supervised signals are the mixture of these cosine functions without noise. And we trained FTNet with the first 800 points ({\color{blue}blue}) and forecast the future 100 points ({\color{red}red}).

The first experiment adapt to the data sets of the \emph{Yancheng Automobile Registration Forecasting} competition \footnote{https://tianchi.aliyun.com/competition/entrance/231641/information} is a real-world univariate time series forecasting task, which requires players to use the daily automobile registration records of a certain period of time (nearly 1000 dates) in the past to predict the number of automobile registration per day for a period of time in the future. Although the actual competition allows the contestant to freely develop other data sets or information as an aid, here we only consider the data of the total number of automobile registration for 5 car brands, not including any specific date information. For simplicity, we fix the test timestamps to be the terminal subset of observations: the last 100 / 1000. This forecasting task is hard since not only objective automobile registration series is the mixture of 5 car brands, but also there exists lots of missing data and sudden changes caused by holiday or other guiding factors which we cannot obtain in advance.

The second experiment uses the \emph{Highway Data of United Kingdom} (HDUK) \footnote{http://data.gov.uk/data set/dft-eng-srn-routes-journey-times}, which is a representative multivariate traffic prediction data set. HDUK contains massive average journey time, speed and traffic flow information for 15-minute periods on all motorways and ``A-level" roads managed by the Highways Agency (known as "the Strategic Road Network" in England). Journey times and speeds are estimated using a combination of sources, including Automatic Number Plate Recognition cameras, in-vehicle Global Positioning Systems and inductive loops built into the road surface. For convenience, we choose roads with relatively large several traffic flow for study and collect the traffic data of the 12 months in 2011 where the first 10 months are divided as the training set and the later 2 months as the testing set. We also set that input (feature vectors) in this experiment is the normalized value (Total Traffic Flow \& Travel Time \& Fused Average Speed \& Link Length) of all observation points in previous 8 time intervals. Output is the prediction value (Total Traffic Flow) in the next time interval of a target observation point for prediction. Empirically, we add the evaluation indicator, \emph{Confusion Accuracy}, which consists of \emph{True Positive Rate} (TPR) and \emph{True Negative Rate} (TNR), to compare the performance of FTNet with other competing methods. 

Pixel-by-pixel MNIST is a popular challenging image recognition data set, which is standard benchmark to test the performance of a learning algorithm. Following a similar setup to~\cite{arjovsky2016,pillow2005}, the handwritten digit image is converted into a pixel-by-pixel spiking train with $T=784$. Standard split of 60,000 training samples and 10,000 testing samples was used with no data augmentation. In general, we also add a convolution filter to the external input signals at each time stamp. The size of the convolution kernel is preset as $2 \times 2$. For classification, all models except SLAYER employ a softmax function and are optimized by a cross-entropy loss. For SNNs, we use the spiking counting strategy, that is, during training, we specify a target of 20 spikes for the true neuron and 5 spikes for each false neuron; while testing, the output class is the one which generates the highest spike count.

\end{document}